%% file: main.tex
\documentclass[10pt,twocolumn,letterpaper]{article}

\usepackage[pagenumbers]{cvpr} 


\definecolor{cvprblue}{rgb}{0.21,0.49,0.74}
\usepackage[pagebackref,breaklinks,colorlinks,allcolors=cvprblue]{hyperref}
\usepackage{amssymb}
\usepackage{graphicx}
\usepackage{cuted}      
\usepackage{caption}
\usepackage{float}
\usepackage{pifont}
\usepackage{multirow}
\usepackage{algorithm}
\usepackage{algpseudocode}
\usepackage[utf8]{inputenc}
\usepackage{colortbl} 
\usepackage{xcolor} 
\usepackage{tcolorbox} 
\usepackage[table]{xcolor}
\usepackage[dvipsnames]{xcolor}
\newcommand{\cmark}{\textcolor{cvprblue}{\ding{51}}}
\newcommand{\xmark}{\textcolor{lightgray}{\ding{55}}}
\usepackage[symbol]{footmisc}

\def\paperID{} 
\def\confName{CVPR}
\def\confYear{2026}

\title{SNOW: Spatio-Temporal Scene Understanding with World Knowledge for Open-World Embodied Reasoning}

\author{
Tin Stribor Sohn$^{1,3\dagger*}$ \hspace{15pt}Maximilian Dillitzer$^{2,3*}$ \hspace{15pt}Jason J. Corso$^{4,5}$ \hspace{15pt}Eric Sax$^1$ \\
$^1$ Karlsruhe Institute of Technology \hspace{15pt}
$^2$ Esslingen University of Applied Sciences \\
$^3$ Dr. Ing. h.c. F. Porsche AG \hspace{15pt}
$^4$ University of Michigan \hspace{15pt}
$^5$ Voxel51 Inc. \\
{\tt\small tin\_stribor.sohn@porsche.de}
}

\begin{document}
\maketitle
\footnotetext[1]{Equal contribution; $^\dagger$Corresponding author.}

\input{01_Chapter/01_abstract}

\input{01_Chapter/02_introduction}
\input{01_Chapter/03_related_work}
\input{01_Chapter/04_snow}

\input{01_Chapter/05_experiments}
\input{01_Chapter/06_conclusion}

{
    \small
    \bibliographystyle{ieeenat_fullname}
    \bibliography{main}
}

\input{01_Chapter/X_Supplementary}

\end{document}

%% file: 01_Chapter/01_abstract.tex
\begin{strip}
    \centering
    \includegraphics[width=\textwidth]{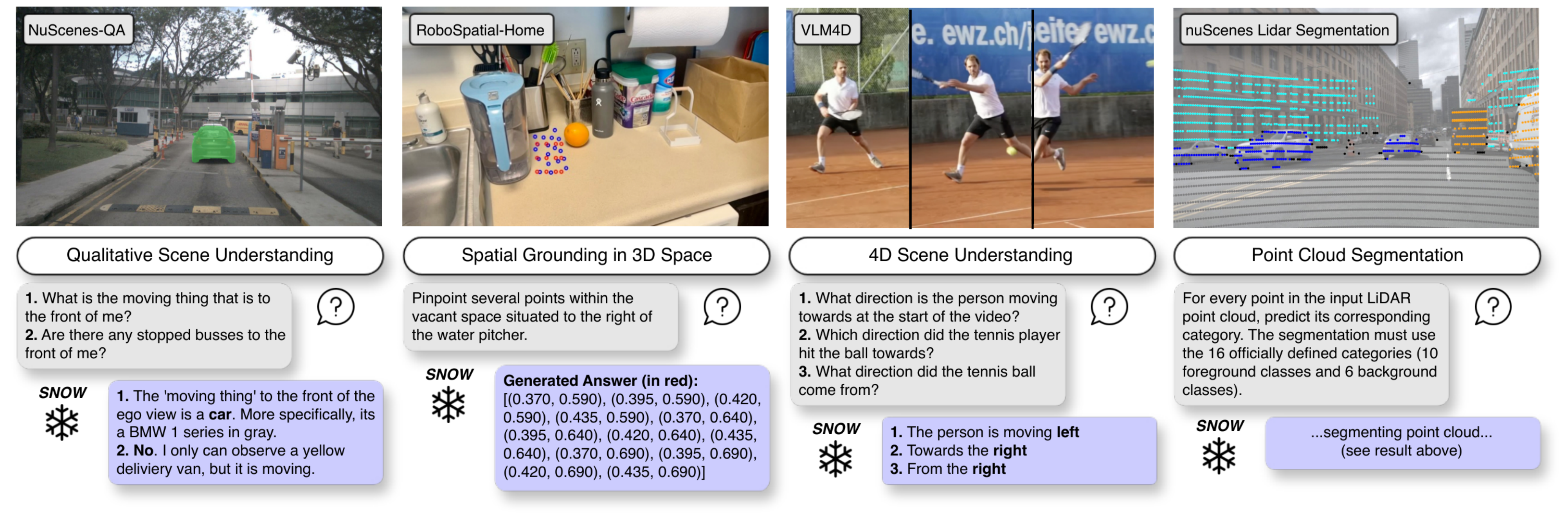}
    \captionof{figure}{\textbf{Overview of SNOW.} SNOW builds a unified 4D Scene Graph (4DSG) by merging VLM semantics with 3D geometry and temporal continuity. STEP tokens encode object-level semantic, spatial, and temporal attributes into a persistent representation that enables grounded reasoning across diverse 4D benchmarks without additional training.}
    \label{fig:firstPage}
\end{strip}

\begin{abstract}
    Autonomous robotic systems require spatio-temporal understanding of dynamic environments to ensure reliable navigation and interaction. While Vision-Language Models (VLMs) provide open-world semantic priors, they lack grounding in 3D geometry and temporal dynamics. Conversely, geometric perception captures structure and motion but remains semantically sparse.
    We propose \textbf{SNOW} (Scene Understanding with Open-World Knowledge), a training-free and backbone-agnostic framework for unified 4D scene understanding that integrates VLM-derived semantics with point cloud geometry and temporal consistency. SNOW processes synchronized RGB images and 3D point clouds, using HDBSCAN clustering to generate object-level proposals that guide SAM2-based segmentation. Each segmented region is encoded through our proposed \textbf{Spatio-Temporal Tokenized Patch Encoding (STEP)}, producing multimodal tokens that capture localized semantic, geometric, and temporal attributes. These tokens are incrementally integrated into a \textbf{4D Scene Graph (4DSG)}, which serves as 4D prior for downstream reasoning.
    A lightweight SLAM backend anchors all STEP tokens spatially in the environment, providing the global reference alignment, and ensuring unambiguous spatial grounding across time. The resulting 4DSG forms a queryable, unified world model through which VLMs can directly interpret spatial scene structure and temporal dynamics. 
    Experiments on a diverse set of benchmarks demonstrate that \textit{SNOW} enables precise 4D scene understanding and spatially grounded inference, thereby setting new state-of-the-art performance in several settings, highlighting the importance of structured 4D priors for embodied reasoning and autonomous robotics.
\end{abstract}

%% file: 01_Chapter/02_introduction.tex
\section{Introduction}
Robotic systems operating in unstructured, dynamic environments must reason not only about which objects are present, but also how they are situated in 3D space and how they evolve over time. This requires the integration of open-world semantics with geometrically precise and temporally coherent scene representations~\cite{Sohn_2025_arXiv_Framework4C}. Existing approaches expose a fundamental disconnect: Vision-Language Models (VLMs), relying on tokenized image patches, provide rich semantic priors and general world knowledge~\cite{Hwang_2024_arXiv_EMMA,Mao_2023_arXiv_GPT-Driver,Xu_2024_IEEE_DriveGPT4}, yet their reasoning remains weakly grounded in spatial geometry and temporal continuity~\cite{Wang_2024_arXiv_LLMSpatial,Zhang_2024_arXiv_LLMSpatial,Ranasinghe_2024_CVPR_LLMSpatial,Chu_2024_arXiv_TemporalReasoning,Wang_2024_arXiv_TemporalReasoning}. Conversely, geometric perception systems capture structure and motion, but are limited in semantic expressiveness and open-vocabulary flexibility. A unified 4D representation within VLMs is therefore required to connect semantic abstraction with persistent spatial and temporal grounding.

To address this challenge, we introduce \textbf{SNOW} (\textbf{S}cene U\textbf{n}derstanding with \textbf{O}pen-\textbf{W}orld Knowledge), a \textit{training-free} framework that constructs a structured 4D representation from synchronized RGB images and point cloud observations. Consecutive point clouds are grouped via HDBSCAN clustering~\cite{Campello_2013_Springer_HDBSCAN} to form object-level proposals, which guide SAM2~\cite{Ravi_2024_arXiv_SAM2} in targeted segmentation. Through calibrated projection and fusion, each segmented region is associated with its geometric shape and temporal identity. We encode each object-level region using our new \textbf{Spatio-Temporal Tokenized Patch Encoding (STEP)}, a compact multimodal token representation capturing localized semantics, geometry, and time.

Accumulating STEP tokens across frames yields a structured \textbf{4D Scene Graph (4DSG)}, where object entities are persistently indexed spatio-temporally by a SLAM backend~\cite{Guadagnino_2025_arXiv_kissSLAM,Keetha_2025_arXiv_MapAnything}. The 4DSG provides a queryable 4D prior: VLMs can infer on spatially grounded semantics and temporally coherent object tracks, without fine-tuning or architectural modification. This representation enables long-horizon reasoning, stable scene interpretation under motion, and consistent integration of new observations.

\noindent The contributions of this work are as follows:
\begin{itemize}
    \item We propose \textbf{SNOW}, a training-free framework that fuses open-world semantic priors from VLMs with temporally consistent 3D perception for 4D scene understanding.
    \item We introduce \textbf{STEP encoding}, a multimodal object-level tokenization scheme that jointly encodes semantic, geometric, and temporal information.
    \item We construct a persistent \textbf{4DSG} that serves as a structured and queryable spatio-temporal representation through which VLMs can perform grounded reasoning in 4D.
    \item We demonstrate that structured 4D priors substantially improve spatial grounding, temporal coherence, and open-vocabulary scene understanding, achieving new state-of-the-art results on multiple benchmarks.
\end{itemize}

\noindent By linking semantic world knowledge of VLMs to explicit 4D structure, SNOW establishes a general foundation for grounded reasoning in autonomous and embodied systems. All code will be open sourced upon publication.

%% file: 01_Chapter/03_related_work.tex
\section{Related Work}
\label{sec:related_work}

Recent advances in VLMs demonstrate strong progress in semantic reasoning and language-guided interaction~\cite{Cui_IEEE_2024_DriveAsYouSay_LLM,Cui_2024_WACV_DriveAsYouSpeak_LLM}. However, spatial and temporal reasoning capabilities remain comparatively underdeveloped~\cite{Wang_2024_arXiv_LLMSpatial,Ranasinghe_2024_CVPR_LLMSpatial,Wang_2024_arXiv_TemporalReasoning,Chu_2024_arXiv_TemporalReasoning}. Spatial reasoning is critical for localization and relational grounding, yet most VLMs either rely on compressed image embeddings or approximate geometric priors~\cite{Hong_2023_arXiv_3dLLM,He_2024_arXiv_UniMOV3D,Mao_2025_arXiv_SpatialLM}. Temporal reasoning, on the other hand, is often reduced to frame-level extensions of image models, where sequential dependencies are captured without maintaining explicit spatial structure~\cite{Li_2024_arXiv_VideoChat_Temporal,Lin_2024_arXiv_VideoLLaVA_Temporal,Li_2024_CVPR_VideoChat2_Temporal}. As a result, existing approaches struggle to jointly represent evolving 3D environments in a manner that is consistent across both space and time. 

\subsection{Spatial Grounding and Representations}
Consider this broad grouping of spatial extensions: image, point cloud, and hybrid modality-based methods~\cite{Zha_2025_arXiv_SurveySpatialLLMs}. Image-based approaches leverage multi-view reconstruction, depth estimation, or Bird’s-Eye-View abstractions to approximate 3D structure~\cite{Zhu_2025_arXiv_LLaVA-3D,Ding_2024_CVPR_BEV-InMLLM,Wei_2024_arXiv_OccLLaMA,Jiao_2025_arXiv_LaVidaDrive,Zhou_2025_ECCV_ELM}. While effective in controlled or dense settings, they often depend on specialized training and pre-aligned feature spaces. Point-based reasoning directly incorporates 3D geometry, either through projection into 2D depth maps~\cite{Zhang_2022_CVPR_PointCLIP,Zhu_2023_ICCVPointCLIPv2} or with dedicated 3D encoders~\cite{Qi_2024_CVPR_GPT4Point,Chen_2024_arXiv_Grounded3DLLM,Deng_2025_arXiv_3D-LLaVA,Zhu_2025_ECCV_ScanReason}. Such methods can achieve fine-grained localization, but typically require multi-stage retraining and are bound to specific modalities. Hybrid designs combine images and point clouds to balance semantic richness with metric fidelity~\cite{Guo_2023_arXiv_Point-LLM,Liu_2024_arXiv_Uni3D-LLM,Ji_2025_IEEE_JM3D,Zhou_2025_arXiv_OpenDriveVLA}, yet this again couples performance to tailored data pipelines and trained backbone models. 

\subsection{Temporal Grounding and Representations}
Temporal extensions have followed a similar trend. Early work uniformly samples video frames and feeds them into pretrained image-language backbones~\cite{Li_2024_arXiv_VideoChat_Temporal,Lin_2024_arXiv_VideoLLaVA_Temporal,Maaz_2024_arXiv_VideoChatGPT_Temporal}, while more advanced models introduce temporal-aware embeddings or memory mechanisms to capture longer sequences~\cite{He_2024_CVPR_MA-LLM_Temporal,Liu_2024_ECCV_ST-LLM_Temporal,Li_2025_ECCV_LLaMA-VID_Temporal}. These approaches improve event localization and activity recognition but treat videos as 2D temporal streams, neglecting the underlying 3D geometry. Consequently, temporal understanding remains largely decoupled from spatial reasoning.  

\subsection{Spatio-Temporal Grounding}
Spatio-temporal grounding attempts to bridge this gap by jointly modeling objects in space and time. Classical computer vision pipelines detect and track spatio-temporal tubes~\cite{Tang_2022_IEEE_HCSTVG,Zhang_2020_CVPR_VidSTG,Yang_2022_CVPR_TubeDETR}, while recent VLM-based variants incorporate temporal encoding into spatial features~\cite{Munasinghe_2023_arXiv_PGVideoLLaVA,Li_2025_CVPR_LLaVAST,Wang_2025_arXiv_SpaceVLLM}. Although these methods demonstrate progress towards dynamic scene understanding, all require task-specific training and remain dependent on fixed backbones. 

\subsection{Gap Towards Unified, Training-free Spatio-Temporal Understanding}
Overall, existing approaches are limited by three factors: (i)~reliance on extensive training to align modalities, (ii)~dependence on specific backbone architectures and model sizes, and (iii)~a lack of explicit geometry when extending into the temporal domain. In particular, alignment-based strategies often sacrifice generalization across modalities and datasets, since optimization is tailored to specific sensory inputs and task settings. This motivates training-free and backbone-agnostic methods that maintain generality, while remaining adaptable to different downstream VLMs. Such approaches must also support diverse point cloud sources (e.g., LiDAR, radar, and RGB-D scans) and preserve spatio-temporal consistency without retraining.

%% file: 01_Chapter/04_snow.tex
\section{Method}
\label{sec:method}

\begin{figure*}[t]
    \centering
    \includegraphics[width=\textwidth]{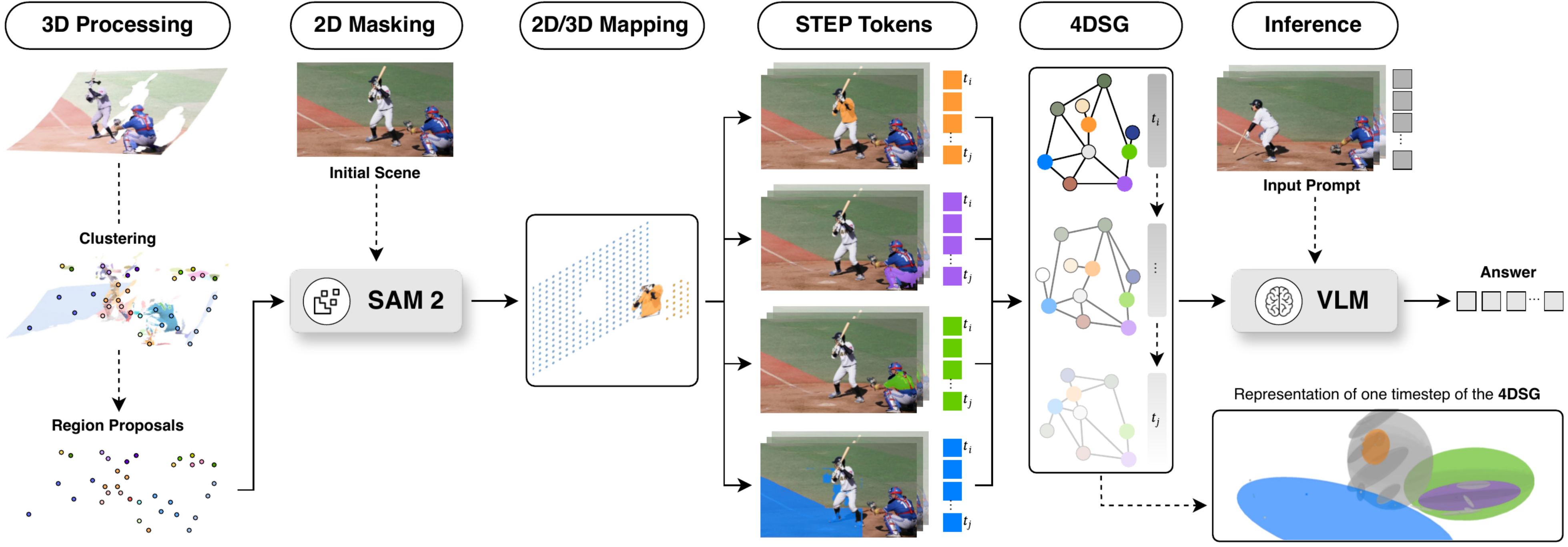} 
    \caption{\textbf{High-level pipeline of SNOW.} The method clusters point clouds, samples representative points, and employs them as point prompts for SAM2-based segmentation. The resulting STEP tokens form a unified spatio-temporal scene graph (i.e., 4DSG), which serves as a persistent 4D world model, queryable by VLMs.}
    \label{fig:SceUnOW_Method}
\end{figure*}

Robotic perception requires the integration of semantic richness, geometric precision, and temporal consistency. While point clouds provide accurate 3D structure, they are semantically sparse. Conversely, VLMs offer open-vocabulary semantics but lack grounding in metric space and temporal reasoning. To bridge this gap, we propose SNOW, a \textit{training-free} and \textit{backbone-agnostic} method for 4D spatio-temporal scene understanding. SNOW operates on synchronized RGB images and point clouds obtained from LiDAR sensors or monocular visual reconstructions via MapAnything~\cite{Keetha_2025_arXiv_MapAnything}. All sensors are assumed to be temporally aligned and geometrically calibrated. The approach leverages 3D point clouds to guide SAM2-based segmentation~\cite{Ravi_2024_arXiv_SAM2}, enforces cross-view and temporal consistency, and organizes all observations into a tokenized 4DSG that serves as a persistent 4D prior to VLMs (cf. Figure~\ref{fig:SceUnOW_Method}). A SLAM backend~\cite{Guadagnino_2025_arXiv_kissSLAM,Keetha_2025_arXiv_MapAnything} is used for maintaining a globally consistent reference frame, ensuring unambiguous spatial alignment. On a single NVIDIA H100 GPU, the pipeline processes about 1.1~frames per second (cf. Appendix~\ref{appendix:runtime}), enabling high-fidelity scene representation in 4D for VLM-based interpretation, scene understanding (i.e., VQA), and downstream tasks such as point cloud segmentation.

\subsection{Point Cloud Clustering and Sampling} 
Given an input point cloud at time $t$, $P^t = \{p_i^t\}_{i=1}^N$ with each $p_i^t \in \mathbb{R}^3$, we initialize the set of unmapped points as $U^t \gets P^t$.
We cluster $U^t$ in metric space using HDBSCAN~\cite{Campello_2013_Springer_HDBSCAN}, which identifies regions of high point density and prunes unstable clusters, producing a set of data-driven spatial clusters: 
\begin{equation}
    \mathcal{R}^t = \{ R_1^t, \dots, R_K^t \}.
\end{equation}
From each cluster $R_k^t$, we uniformly sample $m$ representative points  \( V_k^t = \{ v_{k1}^t, \dots, v_{km}^t \} \subset R_k^t \), which act as region proposals for subsequent mask generation (we use $m=4$ in our experiments).

\subsection{Mask Generation and STEP Encoding} 
All points of the input cloud $P^t$ are first projected into the image plane of camera $c$:
\begin{equation}
    (x_i^{\text{img}}, y_i^{\text{img}}) = \pi(p_i^t, I_c^t),
\end{equation}
where $\pi(\cdot)$ denotes the perspective projection using camera intrinsics and extrinsics.  
Within the same process, the projected region proposals $\{V_k^t\}^{\text{img}}$ are used as point prompts for SAM2~\cite{Ravi_2024_arXiv_SAM2}, which returns object masks
\begin{equation}
    m_{k,c}^t \subset I_c^t .
\end{equation}
Consistency between masks of the same physical object across multiple camera views is enforced via Hungarian matching~\cite{Huhn_1955_HungarianMatching}.  

\begin{figure*}[t]
    \centering
    \includegraphics[width=\textwidth]{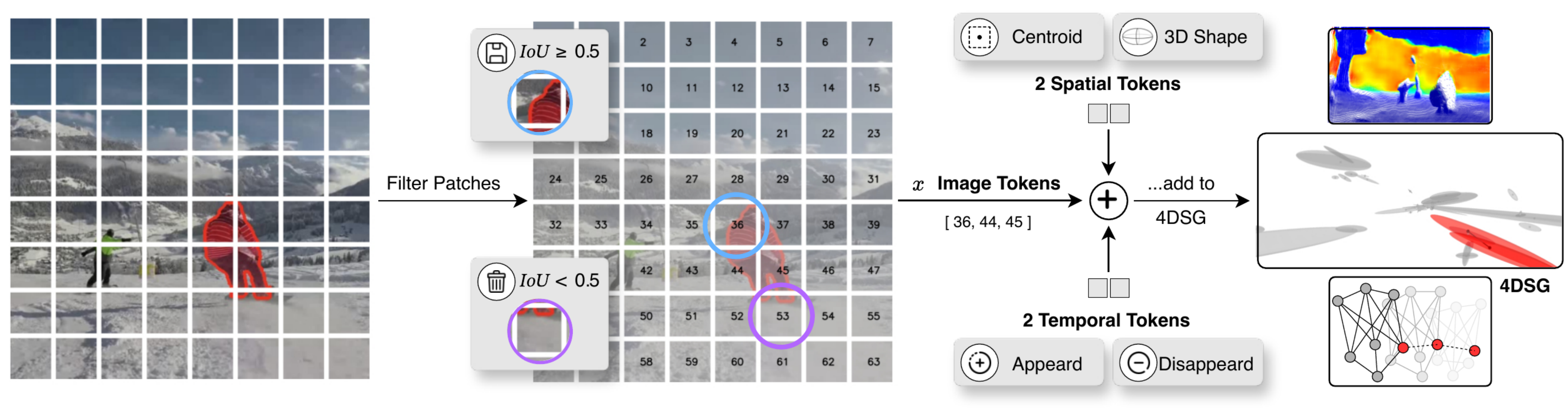}
    \caption{\textbf{STEP token assignment process.} Masks with at least 50\% IoU containment retain their image tokens, which are enriched with 3D centroid, Gaussian shape, and extent tokens, as well as two temporal appearance and disappearance tokens. The resulting STEP tokens are assembled into a 4DSG, serving as SNOW's persistent 4D prior.}
    \label{fig:STEP}
\end{figure*}

Next, we associate the points from the 3D point cloud with their corresponding masks in image space. Each 3D point $(x_i^{\text{img}}, y_i^{\text{img}})$ is assigned to mask $m_{k,c}^t$ if its projection lies within the support of $m_{k,c}^t$ (i.e., $(x_i^{\text{img}}, y_i^{\text{img}}) \in m_{k,c}^t$). Each object mask $m_{k,c}^t$ is passed through our new \textbf{Spatio-Temporal Tokenized Patch Encoding (STEP)}, which compacts semantic, geometric, and temporal information into a unified token representation (cf. Figure~\ref{fig:STEP}). The proposed STEP encoding procedure is as follows:
\begin{enumerate}
    \item The object mask $m_{k,c}^t$ is isolated by coloring all in-mask pixels.
    \item The masked image is partitioned into a fixed $16\times16$ grid, yielding 256 patches.
    \item Each grid cell is evaluated by its Intersection-over-Union (IoU) with the mask. Cells with $\text{IoU} > 0.5$ are retained as \textit{image patch tokens}, denoted ${\tau_{k,1}^t, \dots, \tau_{k,m}^t}$.
    \item To complement the image tokens, four additional feature tokens are appended:
    \begin{itemize}
        \item a \textit{centroid token} $c_k^t = (\bar{x}, \bar{y}, \bar{z})$ encoding the 3D center of the object,
        \item a \textit{shape token} $s_k^t = \big( (\mu_a, \sigma_a, a_{\min}, a_{\max}) \;\big|\; a \in \{x,y,z\} \big)$ derived from Gaussian distributions and spatial extents along each axis, where $\mu_a$ and $\sigma_a$ denote the mean and standard deviation, and $a_{\min}, a_{\max}$ capture the axis-aligned boundaries---this representation preserves the geometric spread of the object without collapsing it into a rectangular bounding box, while simultaneously avoiding skew due to Gaussian approximation and attenuating the influence of outliers through the statistical formulation,
        \item a pair of \textit{temporal tokens} $\theta_k^t = (t_{\text{start}}, t_{\text{end}})$ encoding the time of first appearance and disappearance of the object.
    \end{itemize}
\end{enumerate}
The complete token set for object $k$ at time $t$ is therefore
\begin{equation}
    S_k^t = \{ \tau_{k,1}^t, \dots, \tau_{k,m}^t, c_k^t, s_k^t, \theta_k^t \}.
\end{equation}
These STEP tokens jointly capture semantic appearance (image patches), geometric structure over the whole scene layout (centroid and shape), and temporal context (appearance and disappearance), forming the atomic building blocks of the 4DSG. These feature tokens jointly capture the necessary information for downstream reasoning tasks in 4D, while remaining compact. 

After STEP encoding, the unmapped point set $U^t$ is updated and reprocessed for up to $N_{\text{iter}}$ iterations. In each iteration, residual points are reintroduced into SAM2 for refined mask generation, incrementally integrating previously unassigned structures into the STEP token space. To enhance global consistency, an $H_{\text{hop}}$-step reasoning procedure operates on the tokenized representations, detecting implausible geometries (e.g., elongated Gaussians such as a $50\,\mathrm{m}$ car roof) and reassigning them to $U^t$ (cf. Table~\ref{tab:nhop_refinement_example}). These points are reintegrated into the refinement loop, preventing error accumulation and preserving a consistent spatio-temporal representation.

\subsection{4D Scene Graph Construction}
At each time step $t$, SNOW constructs a spatial scene graph
\begin{equation}
    \mathcal{G}^t = (\mathcal{V}^t, \mathcal{E}^t),
\end{equation}
where each node $v_k^t \in \mathcal{V}^t$ corresponds to a STEP-token set $S_k^t$ representing a localized object instance, and edges $\mathcal{E}^t$ encode spatial relations derived from geometric proximity and relative orientation. This per-frame graph captures the semantic and geometric structure of the scene at a single timestamp.

\paragraph{Spatio-Temporal Association.}
To model temporal evolution, spatial scene graphs are aggregated over a sliding window of $T$ frames,
\begin{equation}
    \mathcal{G}^{t-T:t} = \{\mathcal{G}^{t-T}, \dots, \mathcal{G}^{t}\}.
\end{equation}
Each detected object instance $k$ is associated across frames by using semantic and 3D spatial cues derived from the enriched cluster representation in a STEP token set $S_k^t$. This yields a temporally coherent sequence of STEP tokens for each object
\begin{equation}
    \mathcal{F}_k = \{ S_k^{t-T}, \dots, S_k^{t} \},
\end{equation}
which jointly captures semantic identity, geometric extent, and motion-consistent state progression. Newly observed instances detected in $U^t$ are initialized with fresh STEP tokens, while disappeared ones are terminated by marking their final timestamp $\theta_k^t$. Temporal continuity is therefore encoded directly at the token level, without recurrent state or explicit tracking heuristics. The resulting sequences $\{\mathcal{F}_k\}$ form the node-level temporal representation used in the 4DSG.

\paragraph{4D Scene Graph.}
Aggregating the temporally aligned graphs yields the unified 4DSG
\begin{equation}
    \mathcal{M}^{t} = \bigl(\mathcal{G}^{t-T:t}, \{S_k^{t-T:t}\}\bigr),
\end{equation}
where each object node is represented by a STEP-token sequence encoding semantic attributes, geometric extent, and temporal evolution. To ensure consistent spatial alignment across frames, $\mathcal{M}^t$ is anchored in a globally referenced coordinate system using a SLAM backend: KISS-SLAM~\cite{Guadagnino_2025_arXiv_kissSLAM} for LiDAR input and MapAnything~\cite{Keetha_2025_arXiv_MapAnything} for image-only input. Furthermore, this graph is enriched with the pose and position of the ego actor to provide information needed for self-awareness of the VLM-agent. The resulting 4DSG provides a persistent, queryable 4D prior that unifies spatial and temporal context for downstream reasoning.

The window size $T$ governs the accessible temporal horizon: increasing $T$ enhances long-range reasoning, while decreasing $T$ improves computational efficiency. We use $T = 10$ in our experiments (approximately 10 seconds of context), achieving a balanced trade-off between temporal grounding and inference efficiency. The parameter can be freely adjusted to accommodate different backbone models or resource settings.
By allowing downstream VLMs to operate directly on structured object-level tokens rather than raw sensory data, the 4DSG enables spatially and temporally grounded inference, while remaining training-free.

\subsection{Inference with VLMs}
Inference is performed directly on the unified 4DSG $\mathcal{M}^t$, which provides object-centered semantic and geometric context over the chosen temporal window $T$. Given a query $q$, the VLM reasons on the STEP-token sequences associated with the nodes of $\mathcal{M}^t$, enabling grounding in both recent observations and their temporal evolution.

Formally, we express inference as:
\begin{equation}
    \hat{y} = \text{VLM}(q \mid \mathcal{M}^t),
\end{equation}
where $\mathcal{M}^t$ encodes the spatial layout, semantic attributes, and temporal dynamics of all objects within the window $t\!-\!T:t$, $q$ denotes the user- or task-specific query, and $\hat{y}$ is the resulting output.

Because the 4DSG is already tokenized at the object level, inference requires no additional feature pooling or post-processing. Queries are resolved directly at the object level via STEP-tokens, allowing the VLM to perform grounded reasoning over 4D structure.  Since the representation is training-free and backbone-agnostic, it can be integrated with different VLM architectures and sensing modalities (LiDAR, radar, RGB-D), ensuring flexibility across robotic domains. The full pipeline of SNOW is summarized in Algorithm~\ref{alg:sceunow}, provided in Appendix~\ref{appendix:4dsg_generation}.

\begin{table}[!t]
    \centering
    \small
    \begin{tcolorbox}[colframe=black, colback=gray!10, boxrule=0.25mm, width=\linewidth, title=$H_{\text{hop}}$ Reasoning Refinement Example]
        \textbf{Q:} What object moved the most in the last 2 seconds? \vspace{0.5em}
        
        \textbf{A:} Strictly by world-coordinate displacement between $t=1$ and $t=2$, object\_id 32 (pedestrian) moved the most: about $32.2$ m 
        ($\Delta x \approx +0.13$ m, $\Delta y \approx +32.06$ m, $\Delta z \approx -3.73$ m). This magnitude is implausible for a standing person and the shape/extents changed markedly, suggesting an ID switch or association error. Excluding that outlier and considering consistent tracks, the next largest (and likely true) motion is object\_id 8 (car), which moved about $22.4$ m ($\Delta x \approx -18.61$ m, $\Delta y \approx +12.42$ m, $\Delta z \approx -0.88$ m) over the $2$ s timestep ($\approx11.2$ m/s).
    \end{tcolorbox}
    \caption{\textbf{\(H_{\text{hop}}\) reasoning refinement.} Example of hop-based filtering on the 4DSG, removing implausible motions and geometry outliers before answering spatio-temporal queries.}
    \label{tab:nhop_refinement_example}
\end{table}

%% file: 01_Chapter/05_experiments.tex
\section{Experiments}
\label{sec:experiments}

\subsection{Experimental Setting}
\label{sec:experimentalSetting}

We evaluate SNOW across four complementary benchmarks designed to test semantic, spatial, and temporal understanding. NuScenes-QA~\cite{Qian_2024_AAAI_NuScenes-QA} separately tests spatial and temporal scene comprehension in driving scenes, while RoboSpatial-Home~\cite{Song_2025_arXiv_RoboSpatial} focuses on spatial understanding predominantly. To complement the evaluation setting, we use the VLM4D benchmark~\cite{Zhou_2025_arXiv_VLM4D}, which is designed to assess true 4D understanding of spatial and temporal dynamics in videos. NuScenes-based LiDAR segmentation~\cite{Fong_2021_Lidarseg} provides an additional assessment of spatial accuracy and temporal consistency on a downstream task. An ablation study on VLM4D further isolates the contribution of the integration of 4D STEP tokens in the reasoning process. All evaluations adhere to the official scoring protocols of the respective benchmarks.

For all experiments, we configure SNOW with a local observation window of $T = 10$ frames for temporal tracking, and $N_{\text{iter}} = 1$ and $H_{\text{hop}} = 1$ for refinement, which provides a balance between efficiency and fidelity. The video predictor of SAM2\_Hiera\_Large~\cite{Ravi_2024_arXiv_SAM2} is employed together with Gemma3-4B-IT~\cite{Google_2025_Gemma3} as the backbone VLM. KISS-SLAM~\cite{Guadagnino_2025_arXiv_kissSLAM} serves as the primary SLAM backend, while MapAnything~\cite{Keetha_2025_arXiv_MapAnything} is used for experiments that operate exclusively on image data.

\subsection{Main Results} 
\label{sec:results}

\paragraph{NuScenes-QA Evaluation.}
As shown in Table~\ref{tab:evaluationNuScenesQA}, SNOW establishes a new state-of-the-art on NuScenes-QA~\cite{Qian_2024_AAAI_NuScenes-QA} with an overall accuracy of 60.1\% despite operating entirely without training or fine-tuning. The most pronounced improvement appears in the \textit{Status} category (+23.5\%), indicating that SNOW's STEP-tokenized spatio-temporal 4DSG enables explicit reasoning over dynamic object states such as motion, orientation, or occlusion. Additional gains emerge in \textit{Count} (+4.7\%) and \textit{Object} (+2.9\%), reflecting enhanced multi-entity grounding and robust object identity preservation. Performance in \textit{Existence} and \textit{Comparison} remains comparable to prior work. These results highlight that SNOW leverages multimodal spatio-temporal grounding not only to answer static visual queries but also to integrate evidence across frames, supporting richer 4D scene understanding without domain-specific adaptation.

\begin{table}[t!]
    \centering
    \footnotesize
    \resizebox{\linewidth}{!}{
    \begin{tabular}{l|ccccc|c}
        \toprule
        \textbf{Method}       
        & \textbf{Ext} $\uparrow$ & \textbf{Cnt} $\uparrow$ & \textbf{Obj} $\uparrow$ & \textbf{Sts} $\uparrow$ & \textbf{Cmp} $\uparrow$ & \textbf{Acc} $\uparrow$ \\
        \midrule
        LLaMA-AdapV2~\cite{Gao_2023_arXiv_LLaMA-AdapV2}          & 19.3 & 2.7 & 7.6 & 10.8 & 1.6 & 9.6 \\
        LLaVA1.5~\cite{Liu_2024_CVPR_LLaVA1.5}                   & 45.8 & 7.7 & 7.8 & 9.0 & 52.1 & 26.2 \\
        LiDAR-LLM~\cite{Yang_2023_arXiv_LiDAR-LLM}               & 74.5 & 15.0 & 37.8 & 45.9 & 57.8 & 48.6 \\
        OccLLaMA3.1~\cite{Wei_2024_arXiv_OccLLaMA}               & 80.9 & 19.2 & 46.3 & 47.8 & 66.6 & 54.5 \\
        BEVDet+BUTD~\cite{Qian_2024_AAAI_NuScenes-QA}            & 83.7 & 22.0 & 48.8 & 52.0 & 67.7 & 57.0 \\
        OpenDriveVLA-0.5B~\cite{Zhou_2025_arXiv_OpenDriveVLA}    & 83.9 & 22.0 & 50.2 & \underline{57.0} & 68.4 & 58.4 \\
        OpenDriveVLA-3B~\cite{Zhou_2025_arXiv_OpenDriveVLA}      & \underline{84.0} & 22.3 & \underline{50.3} & 56.9 & \underline{68.5} & \underline{58.5} \\
        OpenDriveVLA-7B~\cite{Zhou_2025_arXiv_OpenDriveVLA}      & \textbf{84.2} & \underline{22.7} & 49.6 & 54.5 & \textbf{68.8} & 58.2 \\
        \midrule
        \textbf{SNOW} (Ours)                                     & 82.3 & \textbf{27.4} & \textbf{53.2} & \textbf{80.5} & 61.0 & \textbf{60.1} \\
        \bottomrule
    \end{tabular}}
    \caption{\textbf{NuScenes-QA evaluation.} Accuracy ($\%$) scores include Existence (Ext), Count (Cnt), Object, (Obj), Status (Sts), Comparison (Cmp), and overall Accuracy (Acc). \textbf{Bold} indicates the highest score and \underline{underline} the second highest score.}
    \label{tab:evaluationNuScenesQA}
\end{table}

\begin{table}[t!]
    \centering
    \footnotesize
    \resizebox{\linewidth}{!}{
    \begin{tabular}{l|cccc}
        \toprule
        \textbf{Method} & \textbf{Cfg.} $\uparrow$ & \textbf{Ctxt.} $\uparrow$ & \textbf{Cpt.} $\uparrow$ & \textbf{Avg.} $\uparrow$ \\
        \midrule
        VILA~\cite{Song_2025_arXiv_RoboSpatial}             & 57.8 &  0.0 & 69.0 & 42.3 \\
        VILA +RS~\cite{Song_2025_arXiv_RoboSpatial}         & 65.9 & 15.6 & 78.0 & 53.2 \\
        LLaVA-NeXT~\cite{Song_2025_arXiv_RoboSpatial}       & 68.3 &  0.0 & 70.5 & 46.3 \\
        LLaVA-NeXT +RS~\cite{Song_2025_arXiv_RoboSpatial}   & \underline{78.9} & 19.7 & \underline{80.1} & 59.6 \\
        SpaceLLaVA~\cite{Song_2025_arXiv_RoboSpatial}       & 61.0 &  2.5 & 61.0 & 41.5 \\
        SpaceLLaVA +RS~\cite{Song_2025_arXiv_RoboSpatial}   & 71.6 & 13.1 & 72.4 & 52.4 \\
        RoboPoint~\cite{Song_2025_arXiv_RoboSpatial}        & 69.9 & 19.7 & 70.5 & 53.4 \\
        RoboPoint +RS~\cite{Song_2025_arXiv_RoboSpatial}    & 78.0 & \underline{31.1} & \textbf{81.0} & \underline{63.4} \\
        3D-LLM~\cite{Song_2025_arXiv_RoboSpatial}           & 39.8 &  0.0 & 35.2 & 25.0 \\
        3D-LLM +RS~\cite{Song_2025_arXiv_RoboSpatial}       & 55.2 &  8.2 & 52.3 & 37.6 \\
        LEO~\cite{Song_2025_arXiv_RoboSpatial}              & 51.2 &  0.0 & 38.1 & 29.8 \\
        LEO +RS~\cite{Song_2025_arXiv_RoboSpatial}          & 64.2 & 10.0 & 57.1 & 43.8 \\
        Molmo~\cite{Song_2025_arXiv_RoboSpatial}            & 58.6 &  0.1 & 18.1 & 25.6 \\
        GPT-4o~\cite{Song_2025_arXiv_RoboSpatial}           & 77.2 &  5.7 & 58.1 & 47.0 \\
        NaviMaster~\cite{Luo_2025_arXiv_NaviMaster}         & -- & 21.65 & -- & -- \\
        \midrule
        \textbf{SNOW} (Ours)                                & \textbf{84.55} & \textbf{54.92} & 78.10 & \textbf{72.29} \\
        \bottomrule
    \end{tabular}}
    \caption{\textbf{RoboSpatial-Home grounded VQA evaluation.} Models are evaluated on three dimensions: Configuration (Cfg.), Context (Ctxt.), and Compatibility (Cpt.), with Avg. reporting their mean. ``+RS" denotes models finetuned on the RoboSpatial dataset. SNOW operates in a fully training-free setting, using its 4DSG as a persistent spatial representation. \textbf{Bold} indicates the highest score and \underline{underline} the second highest score.}
    \label{tab:robospatial_results}
\end{table}

\begin{table*}[!t]
    \centering
    \footnotesize
    \begin{tabular}{l|ccc|ccc|c}
        \toprule
        \textbf{Model} & \textbf{Ego-C.} $\uparrow$ & \textbf{Exo-C.} $\uparrow$ & \textbf{Avg.} $\uparrow$ & \textbf{Direct.} $\uparrow$ & \textbf{FP} $\uparrow$ & \textbf{Avg.} $\uparrow$ & \textbf{Overall} $\uparrow$ \\
        \midrule
        GPT-4o~\cite{Zhou_2025_arXiv_VLM4D}                   & 55.5 & 62.2 & 60.0 & 49.5 & 53.3 & 49.9 & 57.5 \\
        Gemini-2.5-Pro~\cite{Zhou_2025_arXiv_VLM4D}           & \underline{64.6} & \underline{62.9} & \underline{63.5} & \underline{54.8} & \underline{80.0} & \underline{57.3} & \underline{62.0} \\
        Claude-Sonnet-4~\cite{Zhou_2025_arXiv_VLM4D}          & 52.6 & 52.1 & 52.2 & 44.0 & \textbf{86.7} & 48.3 & 51.3 \\
        Llama-4-Maverick-17B~\cite{Zhou_2025_arXiv_VLM4D}     & 52.6 & 54.3 & 53.8 & 53.3 & 51.1 & 53.0 & 53.6 \\
        Llama-4-Scout-17B~\cite{Zhou_2025_arXiv_VLM4D}        & 48.6 & 56.2 & 53.7 & 53.3 & 75.6 & 55.5 & 54.1 \\
        Qwen2.5-VL-72B~\cite{Zhou_2025_arXiv_VLM4D}           & 54.3 & 52.5 & 53.1 & 49.5 & \underline{80.0} & 52.6 & 53.0 \\
        InternVideo2.5-8B~\cite{Zhou_2025_arXiv_VLM4D}        & 57.2 & 50.5 & 52.7 & 44.3 & 46.7 & 44.5 & 50.7 \\
        \midrule
        \textbf{SNOW} (Ours)                                  & \textbf{73.04} & \textbf{72.78} & \textbf{72.87} & \textbf{71.16} & 77.86 & \textbf{76.46} & \textbf{73.75} \\
        \bottomrule
    \end{tabular}
    \caption{\textbf{VLM4D evaluation.} Accuracy ($\uparrow$) is reported for egocentric (Ego-C.) and exocentric (Exo-C.) reasoning, their average, directional (Direct.), and false positive reasoning (FP). The final columns provide the average across reasoning types and the overall benchmark score. \textbf{Bold} indicates the highest score and \underline{underline} the second highest score.}
    \label{tab:vlm_eval}
\end{table*}

\paragraph{RoboSpatial-Home Evaluation.}
RoboSpatial-Home~\cite{Song_2025_arXiv_RoboSpatial} evaluates grounded spatial reasoning in real indoor environments. The benchmark tests three complementary dimensions of spatial understanding: (i) \textit{Spatial Context} measures whether a model can identify suitable free or support surfaces by predicting a point location in the scene; (ii) \textit{Spatial Compatibility} evaluates whether a region can feasibly support a given object, formulated as binary feasibility judgments; and (iii) \textit{Spatial Configuration} assesses relative object-to-object spatial relationships. 

\begin{figure}[!t]
    \centering
    \begin{subfigure}{\linewidth} 
        \centering
        \includegraphics[width=0.48\linewidth]{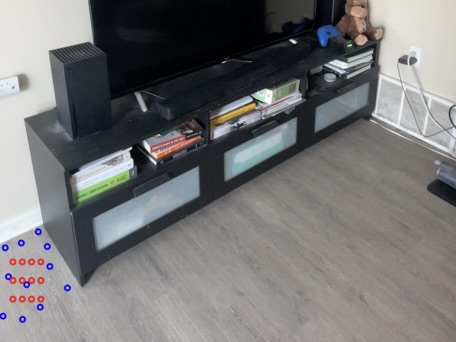}
        \includegraphics[width=0.48\linewidth]{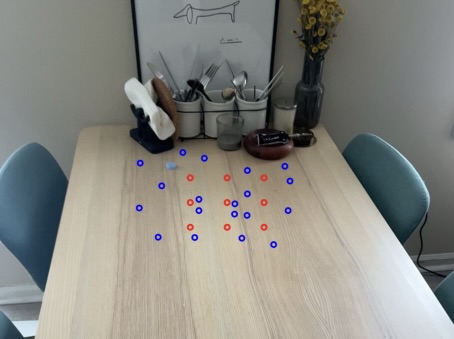}
    \end{subfigure}%
    \vspace{0.5em}
    \begin{subfigure}{\linewidth} 
        \centering
        \includegraphics[width=0.48\linewidth]{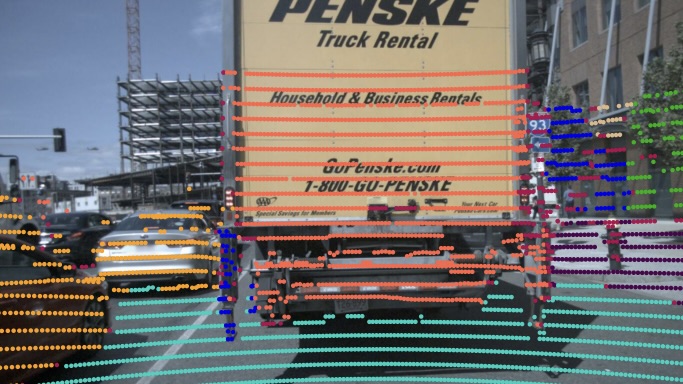}
        \includegraphics[width=0.48\linewidth]{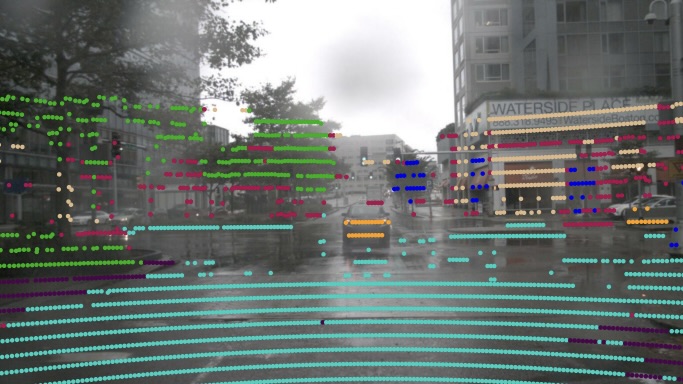}
    \end{subfigure}%
    \caption{\textbf{Qualitative examples of SNOW on RoboSpatial-Home and open-vocabulary LiDAR segmentation.} For RoboSpatial-Home, \textcolor{red}{red} denotes the model prediction; \textcolor{blue}{blue} denotes the ground truth reference.}
    \label{fig:robospatial_lidarseg}
\end{figure}

We compare SNOW against pretrained and RoboSpatial-finetuned models (cf. Table~\ref{tab:robospatial_results}). Unlike approaches requiring task-specific finetuning or spatial alignment training, SNOW performs zero-shot grounding via its STEP-based 4DSG.
SNOW establishes a new state-of-the-art average performance on RoboSpatial-Home of 72.29\% (cf. Table~\ref{tab:robospatial_results}). Most notably, SNOW improves \textit{Spatial Context} by a substantial margin of +23.82\%, the most challenging dimension requiring continuous-point spatial grounding, outperforming all prior systems including those explicitly finetuned on RoboSpatial (``+RS") (cf. Figure~\ref{fig:robospatial_lidarseg}). SNOW further achieves +7.35\% in \textit{Spatial Configuration} and remains competitive in \textit{Spatial Compatibility} (-2.9\%), indicating consistent generalization across complementary spatial reasoning tasks. Critically, SNOW attains these results without training, whereas prior leading approaches rely on benchmark-specific finetuning and spatial alignment training. These findings demonstrate that structured 4D scene representations and STEP-based grounding enable strong spatial understanding. Further qualitative success and failure cases are provided and discussed in Appendix~\ref{appendix:robospatial}.

\paragraph{VLM4D Evaluation.}
Table~\ref{tab:vlm_eval} presents a comprehensive comparison of SNOW against state-of-the-art models on the VLM4D benchmark. SNOW achieves 73.04\% ego-centric and 72.78\% exo-centric reasoning accuracy, corresponding to absolute improvements of +8.44\% and +9.88\% over the strongest baseline (Gemini-2.5-Pro), and yields an average reasoning gain of +9.37\%. In directional reasoning, SNOW attains 71.16\%, surpassing the best prior model by a significant margin of +16.36\%.  

For false positive (FP) reasoning, SNOW scores 77.86\%, which is comparable to high-performing baselines, confirming that 4D spatio-temporal modeling is primarily beneficial for scenario comprehension rather than for FP detection. Overall, SNOW achieves an overall benchmark score of 73.75\%, outperforming all baselines substantially by +11.75\% on average. These results quantitatively underline that the integration of 4D STEP tokens significantly enhances the model's ability to reason about space and time, particularly in ego-, exo-centric, and directional contexts. Qualitative examples are provided in Appendix~\ref{appendix:vlm4d} to further illustrate SNOW's spatio-temporal understanding along with success and failure cases.

\paragraph{Downstream Tasks Evaluation on NuScenes.}

Table~\ref{tab:evaluationLidarSegmentation} presents a comparison between SNOW and recent open-vocabulary LiDAR segmentation models on NuScenes LiDAR segmentation~\cite{Fong_2021_Lidarseg}. Evaluation is conducted on the \textit{validation split} using the mean IoU (mIoU) metric. Unlike prior methods that depend on task-specific finetuning or adaptation, SNOW performs 3D point-level grounding in a fully training-free manner by directly projecting STEP token embeddings into the LiDAR space. Despite this zero-shot setting, SNOW achieves an mIoU of 38.1, ranking second overall and surpassing several approaches requiring additional training. This result highlights the effectiveness of SNOW's 4D STEP representation, where structured spatial-temporal object embeddings inherently encode transferable geometric and semantic priors, enabling consistent and modality-agnostic instance grounding in 3D scenes. Further qualitative examples can be observed in Figure~\ref{fig:robospatial_lidarseg} and Appendix~\ref{appendix:nuscens}.

\begin{table}[t!]
    \centering
    \footnotesize
    \begin{tabular}{l|cc}
            \toprule
            \textbf{Method} & \textbf{mIoU} $\uparrow$ & \textbf{TF} \\
            \midrule
            CNS~\cite{Chen_2023_arXiv_CNS}                      & 26.8 & \xmark \\
            AdaCo~\cite{Zou_2024_arXiv_AdaCo}                   & 31.2 & \xmark\\
            3D-AVS~\cite{Wei_2025_CVPR_3dAVS}                   & 36.2 & \xmark \\
            OpenScene~\cite{Peng_2023_arXiv_OpenScene}          & 36.7 & \xmark \\
            OV3D~\cite{Jiang_2024_CVPR_OV3D}                    & \textbf{44.6} & \xmark \\
            \midrule
            \textbf{SNOW} (Ours)                                & \underline{38.1} & \cmark \\
            \bottomrule
        \end{tabular}
    \caption{\textbf{LiDAR segmentation.} Comparison of SNOW with open-vocabulary segmentation models on the NuScenes LiDAR segmentation task, using the official mIoU metric. ``TF" indicates weather the method is training-free (\cmark). \textbf{Bold} indicates the highest score and \underline{underline} the second highest score.}
    \label{tab:evaluationLidarSegmentation}
\end{table}

\begin{table*}[!t]
    \centering
    \footnotesize
    \begin{tabular}{l|cc|ccc|ccc|c}
        \toprule
        \textbf{ID} & \textbf{2D + t} & \textbf{4D-STEP} & \textbf{Ego-C.} $\uparrow$ & \textbf{Exo-C.} $\uparrow$ & \textbf{Avg.} $\uparrow$ & \textbf{Drct.} $\uparrow$ & \textbf{FP} $\uparrow$ & \textbf{Avg.} $\uparrow$ & \textbf{Overall} $\uparrow$ \\
        \midrule
        A1$^\dagger$ & \xmark & \xmark  & 38.0 & 42.0 & 40.0 & 43.0 & \textbf{74.0} & 58.5 & 49.25 \\
        A2$^\dagger$ & \cmark & \xmark  & \underline{56.0} & \underline{62.0} & \underline{59.0} & \underline{58.0} & \underline{72.0} & \underline{65.0} & \underline{62.0} \\
        A3$^\dagger$ & \xmark & \cmark  & \textbf{78.0} & \textbf{82.0} & \textbf{80.0} & \textbf{76.0} & \textbf{74.0} & \textbf{75.0} & \textbf{77.5} \\
        \bottomrule
    \end{tabular}
    \caption{\textbf{Ablation study of SNOW on VLM4D.} Each configuration isolates the contributions of temporal linking across frames (``2D + t"), and the 4D STEP tokens (``4D-STEP") over the baseline model. Performance is reported using VLM4D accuracy metrics ($\uparrow$). $^\dagger$~Results are evaluated on a 200 question subset of the benchmark. \textbf{Bold} indicates the highest score and \underline{underline} the second highest score.}    
    \label{tab:ablation}
\end{table*}

\subsection{Ablation Study}
To complement benchmark results, we analyze the contribution of SNOW's core representation components on 4D reasoning performance. All variants are evaluated on a 200 question subset of the VLM4D benchmark~\cite{Zhou_2025_arXiv_VLM4D}, using the same VLM backbone and input time window of $T=10$ frames. Questions are equally distributed across categories (i.e., 50 per benchmark category).

\textbf{A1 (VLM-Only Baseline).}
The backbone model (Gemma3-4B-IT~\cite{Google_2025_Gemma3}) receives RGB frames over the temporal window but no structured multi-view or temporal association. This setting isolates the language and perception capabilities of the VLM without explicit scene structure.

\textbf{A2 (2D Temporal Tracking Only).}
Object instances are tracked over time in the image plane. STEP token appearance and disappearance timestamps are maintained, but no 3D spatial tokens are available. This setting captures temporal continuity but lacks spatial coherence.

\textbf{A3 (Full 4D STEP Representation).}
3D spatial structure and temporal instance links are fused into unified STEP tokens. Each object maintains a temporally indexed sequence of spatially consistent embeddings, forming the basis of the 4DSG representation used by SNOW.

The progression from A1 to A3 highlights the role of structured 4D scene representation in supporting robust reasoning. Introducing only 2D temporal tracking (A2) yields a substantial improvement over the VLM-only baseline (A1), particularly in ego- (+18\%) and exo-centric spatial reasoning (+20\%). This indicates that maintaining object identity across time is already a strong inductive prior for understanding dynamic scenes. However, without spatial grounding, reasoning remains limited when queries involve viewpoint transformation or require resolving object interactions in 3D space. FP does not benefit from the 4D STEP representation, as these questions do not require spatial or temporal grounding but solely the reasoning whether objects are present or not, which is given in the patch tokens already provided in the baseline models.

The full 4D STEP representation (A3) further improves performance across all metrics, most prominently in spatial reasoning (Ego-C., Exo-C.) where we observe gains of +22\% and +20\% over A2. STEP tokens provide temporally indexed 3D-consistent embeddings, enabling SNOW to localize, compare, and relate objects across space-time rather than relying solely on image-plane continuity. This reduces perspective ambiguity and allows the model to answer queries in 4D involving object placement, motion trajectories, and cross-frame relational constraints. The final ``Overall" score increases from 49.25\% (A1) to 62.0\% (A2) and further to 77.5\% (A3), confirming that 4D spatial-temporal grounding (i.e., STEP tokens) is the dominant contributor to SNOW's reasoning capability.

%% file: 01_Chapter/06_conclusion.tex
\section{Conclusion}

We presented \textbf{SNOW}, a training-free and backbone-agnostic framework for 4D spatio-temporal scene understanding in open-world robotic environments. By clustering point clouds, point-prompting SAM2 for segmentation, and enriching objects with geometric and temporal attributes, SNOW unifies 3D structure, open-vocabulary semantics, and temporal dynamics into a single coherent representation. Its tokenized 4DSG enables compact yet expressive encoding of object-level information and maintains temporal continuity through globally aligned semantic information. This design provides several advantages: (i)~it supports plug-and-play integration with diverse VLMs and sensing modalities, (ii)~generalizes across both static and dynamic environments without retraining, and (iii)~offers persistent memory for long-horizon reasoning and spatio-temporal grounding. SNOW achieves consistent improvements across 4D understanding benchmarks, demonstrating that structured STEP tokenization can serve as a universal interface between geometric perception and foundation models. Beyond perception, SNOW offers a scalable foundation for embodied agents, enabling unified scene interpretation, semantic mapping, and temporal abstraction in physically grounded world models.

\noindent\textbf{Limitations and Future Work.}  
The current implementation accumulates long sequences of STEP tokens, which slows inference on large-scale scenes and long temporal sequences. Also, the 4DSG effectively captures global motion but may underrepresent fine-grained dynamics and object morphing. Future work will therefore explore (i)~explicit point tracking for local motion modeling, (ii)~latent-space fusion modules for faster and more compact token integration, (iii)~encoders with learned 4D representations and attention mechanisms, whereas SNOW could serve as a training pipeline for data acquisition, and~(iv) studies on STEP token ordering and temporal compression to enhance downstream performance. These directions aim to extend SNOW towards scalable, real-time 4D scene understanding.

%% file: 01_Chapter/X_Supplementary.tex
\clearpage
\setcounter{page}{1}
\maketitlesupplementary

\section{4DSG Generation of SNOW} 
\label{appendix:4dsg_generation}

Algorithm~\ref{alg:sceunow} summarizes the full procedure used to construct the 4D Scene Graph (4DSG) described in Section~\ref{sec:method}. The pipeline processes synchronized point clouds and images in a streaming fashion, progressively forming a temporally grounded representation of object-level structure and motion.

At each timestep, the input point cloud is iteratively partitioned into object-level regions through a cycle of geometric clustering, multi-view projection, and segmentation refinement using SAM2~\cite{Ravi_2024_arXiv_SAM2}. This iterative formulation serves two purposes: (i)~it progressively resolves object boundaries in challenging cluttered or partially visible scenes, and (ii)~it prevents premature object consolidation by deferring assignment for geometrically implausible clusters. Each stabilized region is encoded into a \textbf{STEP token}, which captures shape, trajectory-consistent position, estimated extent, and appearance or disappearance across time. These tokens form a compact latent representation that supports direct interfacing with VLMs.

To model temporal continuity, STEP tokens associated with the same physical object are linked over a sliding window of $T$ frames. This token-level temporal linking avoids explicit tracking heuristics and ensures that changes in geometry and viewpoint are absorbed naturally into the representation. The resulting temporally aligned STEP sequences form the node embeddings of the 4DSG. Graph edges encode spatial relations derived from 3D proximity and relative orientation.

A SLAM backend maintains a globally consistent coordinate frame, allowing object identities and their spatial positions to remain stable across time. Additionally, they provide ego position and poses over time, accounting for camera motion and embodied agent self-awareness, which both is embedded into the 4DSG. We use KISS-SLAM~\cite{Guadagnino_2025_arXiv_kissSLAM} when LiDAR is available and MapAnything~\cite{Keetha_2025_arXiv_MapAnything} for image-only reconstruction. This ensures that the 4DSG encodes spatial layout and temporal evolution in a common reference frame independent of sensing modality.

The final 4DSG at time $t$ is a queryable object-centric memory of the scene over the temporal window $t\!-\!T:t$. Downstream inference tasks (e.g., open-vocabulary scene understanding, spatio-temporal reasoning) interact directly with the 4DSG, allowing VLMs to operate over structured 4D context instead of raw sensor data. Because the representation is token-based, no additional pooling, feature alignment, or task-specific training is required. The pseudocode in Algorithm~\ref{alg:sceunow} outlines this process step-by-step, illustrating how structured 4D representations emerge from multimodal association and temporal consolidation.

\begin{algorithm}[!t]
\caption{4D Spatio-Temporal Scene Understanding with SNOW and STEP Encoding}
\label{alg:sceunow}
\small
\begin{algorithmic}[1]
\Require Point clouds $\{P^t\}_{0:T}$, image sequence $\{I_c^t\}_{0:T}$, temporal window $T$, iterations $N_{\text{iter}}$, reasoning hops $H_{\text{hop}}$, VLM backbone
\State Initialize persistent 4DSG $\mathcal{M}^0 \gets \emptyset$
\For{each time step $t$}
    \State Initialize unmapped points $U^t \gets P^t$
    \For{$n = 1 \dots N_{\text{iter}}$}
        \State Cluster $U^t \!\to\! \mathcal{R}^t$, sample proposals $V_k^t$
        \State Project all $p_i^t \in P_k^t$ to images $I_c^t$
        \State Prompt SAM2 with $\{V_k^t\}^{\text{img}} \rightarrow$ masks $m_{k,c}^t$
        \State Match across views $\to m_k^t$, assign points $\to \hat{R}_k^t$
        \State Encode objects $\to$ STEP tokens $S_k^t$
        \For{$h = 1 \dots H_{\text{hop}}$}
            \State Detect implausible geometries, reassign to $U^t$
        \EndFor
        \State $U^t \gets P^t \setminus \bigcup_k \hat{R}_k^t$ \If{$U^t=\emptyset$} \textbf{break} \EndIf
    \EndFor
    \State Build spatial scene graph at $t$: $\mathcal{G}^t=(\mathcal{V}^t,\mathcal{E}^t)$
    \State Update temporal representation $\mathcal{F}_k \gets \{S_k^{t-T},\dots,S_k^t\}$
    \State Based on $\mathcal{F}_k$, fuse $\mathcal{G}^t$ into 4DSG $\mathcal{M}^t$
    \State Query VLM with $(q \mid \mathcal{M}^t) \to \hat{y}$
\EndFor
\end{algorithmic}
\end{algorithm}

\section{Further Results on RoboSpatial-Home} 
\label{appendix:robospatial}

\begin{figure*}[!t]
    \centering
    \begin{tabular}{ccc}
    \begin{subfigure}{0.31\linewidth}
        \centering
        \includegraphics[width=\linewidth]{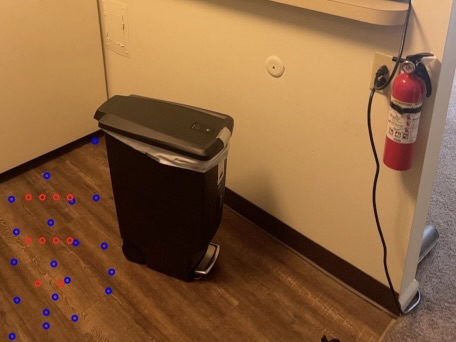}
        \caption{Q.033}
    \end{subfigure} &
    \begin{subfigure}{0.31\linewidth}
        \centering
        \includegraphics[width=\linewidth]{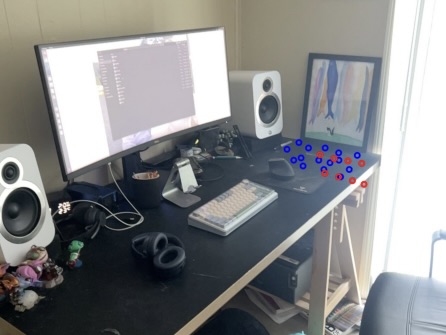}
        \caption{Q.078}
    \end{subfigure} &
    \begin{subfigure}{0.31\linewidth}
        \centering
        \includegraphics[width=\linewidth]{02_Figures/Q97.jpeg}
        \caption{Q.097}
    \end{subfigure} \\
    
    \begin{subfigure}{0.31\linewidth}
        \centering
        \includegraphics[width=\linewidth]{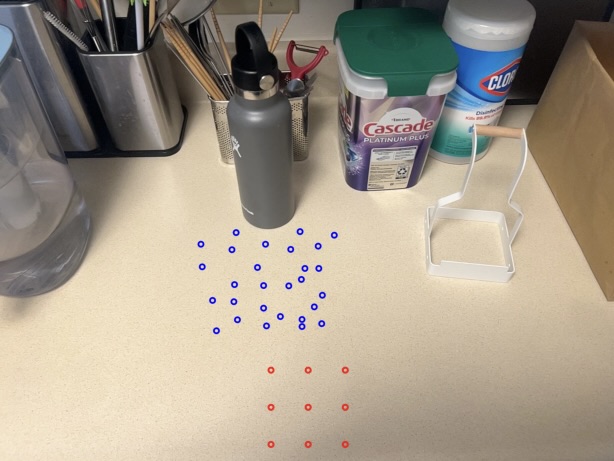}
        \caption{Q.052}
    \end{subfigure} &
    \begin{subfigure}{0.31\linewidth}
        \centering
        \includegraphics[width=\linewidth]{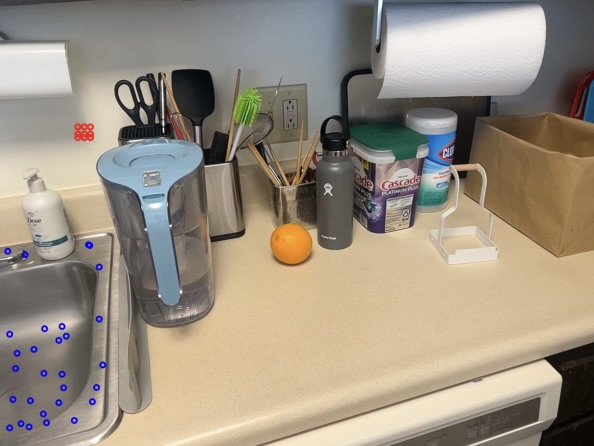}
        \caption{Q.055}
    \end{subfigure} &
    \begin{subfigure}{0.31\linewidth}
        \centering
        \includegraphics[width=\linewidth]{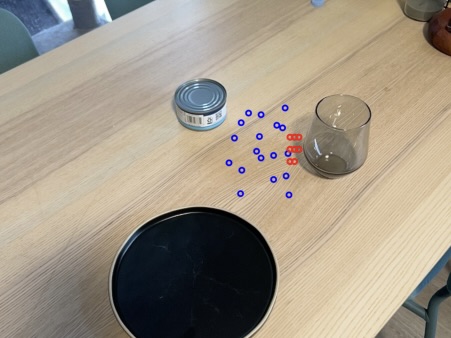}
        \caption{Q.100}
    \end{subfigure} \\
    
    \begin{subfigure}{0.31\linewidth}
        \centering
        \includegraphics[width=\linewidth]{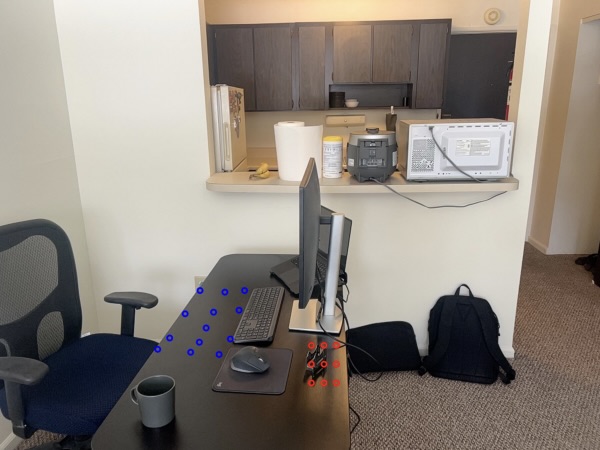}
        \caption{Q.013}
    \end{subfigure} &
    \begin{subfigure}{0.31\linewidth}
        \centering
        \includegraphics[width=\linewidth]{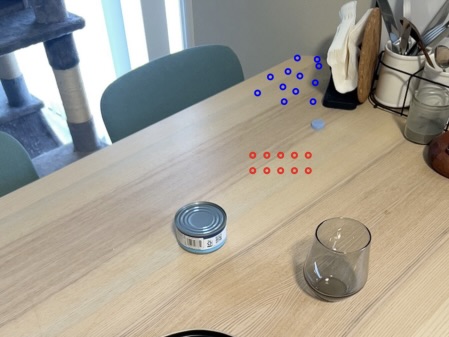}
        \caption{Q.101}
    \end{subfigure} &
    \begin{subfigure}{0.31\linewidth}
        \centering
        \includegraphics[width=\linewidth]{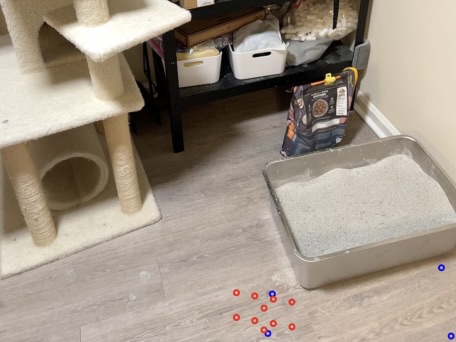}
        \caption{Q.109}
    \end{subfigure}
    \end{tabular}
    \caption{\textbf{Qualitative examples of SNOW on RoboSpatial-Home} illustrating correct predictions (top row), ambiguous cases (middle row), and failure modes (bottom row). \textcolor{red}{Red} denotes the model prediction; \textcolor{blue}{Blue} denotes the ground truth reference.}
    \label{fig:robospatial_appendix}
\end{figure*}

Figure~\ref{fig:robospatial_appendix} presents qualitative examples of SNOW on the RoboSpatial-Home benchmark~\cite{Song_2025_arXiv_RoboSpatial}, focusing on the \textit{pinpointing} task. We highlight this task as it constitutes the most demanding spatial reasoning setting in RoboSpatial-Home. Despite being a training-free approach, SNOW achieves new state-of-the-art performance (cf. Section~\ref{sec:results}), demonstrating strong zero-shot spatial localization and grounding ability.

The first row of Figure~\ref{fig:robospatial_appendix} illustrates representative success cases, in which SNOW accurately infers and pinpoints positions in spatial relation to the referenced object. The middle row shows failure cases that arise not from model limitations but from inherent question ambiguity. For example, several benchmark questions describe spatial relations imprecisely (e.g., ``vacant space in front of the bottle", Q.052, cf. Table~\ref{tab:robospatial_appendix}), where multiple positions are semantically valid and SNOW points to one of those locations. In such cases, SNOW's predictions are reasonable yet counted as incorrect due to the benchmark's single ground-truth polygon annotation. For clarity, we provide the original benchmark question formulations in Table~\ref{tab:robospatial_appendix}. These instances suggest that evaluation errors may originate from low-spec or under-specified spatial language instructions rather than model failure; thus, they should be interpreted cautiously.

The bottom row contains genuine error cases, where SNOW's prediction diverges from the intended spatial relation. Even here, some errors occur when SNOW predicts a region that is spatially correct but lies slightly outside the annotated ground-truth polygon (e.g., Q.55). This reflects a known limitation of polygon-based evaluation for open spatial reasoning tasks, where the ``correct region" is itself continuous rather than discretely bounded.

Overall, SNOW demonstrates robust and generalizable spatial reasoning in the zero-shot setting, but these examples highlight that future benchmarks would benefit from (i)~more precise spatial language phrasing and (ii)~tolerance-based or region-proposal-based evaluation metrics to avoid penalizing semantically valid predictions. We emphasize that these observations are intended to contextualize evaluation behavior rather than critique the benchmark design itself.

\begin{table*}[!t]
    \small
    \centering
    \begin{tabular}{c|l}
        \toprule
         \textbf{ID} & \textbf{RoboSpatial-Home Question} \\
         \midrule
         Q.033 & In the image, there is a fridge. Pinpoint several points within the vacant space situated to the in front of the fridge. \\
         Q.078 & In the image, there is a painting. Pinpoint several points within the vacant space situated to the in front of the painting. \\
         Q.097 & In the image, there is a painting. Pinpoint several points within the vacant space situated to the in front of the painting. \\
         \midrule
         Q.052 & In the image, there is a bottle. Pinpoint several points within the vacant space situated to the in front of the bottle. \\
         Q.055 & In the image, there is a sink. Pinpoint several points within the vacant space situated to the above the sink. \\
         Q.100 & In the image, there is a cup. Pinpoint several points within the vacant space situated to the left of the cup. \\
         \midrule
         Q.013 & In the image, there is a monitor. Pinpoint several points within the vacant space situated to the in front of the monitor. \\
         Q.101 & In the image, there is a tissue. Pinpoint several points within the vacant space situated to the left of the tissue. \\
         Q.109 & In the image, there is a litter box. Pinpoint several points within the vacant space situated to the in front of the litter box. \\
         \bottomrule
    \end{tabular}
    \caption{\textbf{RoboSpatial-Home question formulations} for the spatial pinpointing task examples shown in Figure~\ref{fig:robospatial_appendix}. We provide these to clarify success, failure, and cases where ambiguity in spatial phrasing may influence evaluation outcomes.}
    \label{tab:robospatial_appendix}
\end{table*}

\section{Further Results on VLM4D} 
\label{appendix:vlm4d}

Table~\ref{tab:vlm4d_multiframe_examples} presents selected examples from the VLM4D benchmark to illustrate SNOW's qualitative performance across diverse scenarios. We include representative questions from four categories: \textit{Curling}, \textit{Burnout}, \textit{Desk}, and \textit{Futuristic Car}.  

In the \textit{Curling} and \textit{Desk} scenarios, SNOW perfectly reproduces the ground truth, demonstrating precise ego-centric and exo-centric spatial reasoning, as well as fine-grained action understanding. The \textit{Burnout} scenario highlights more challenging directional reasoning under complex motion; while SNOW occasionally differs from ground truth (e.g., Q.77--Q.79), the model still captures essential scene dynamics, reflecting the limits of purely visual cues without additional context. In the \textit{Futuristic Car} scenario, SNOW correctly identifies static and absent entities, showing robust scene parsing even under occlusion or missing objects.  

Overall, these qualitative examples confirm that 4D STEP token integration enables SNOW to track actors and temporal interactions reliably, providing a strong foundation for reasoning about space, motion, and time in complex 4D environments.

\begin{figure*}[!t]
    \centering
    \begin{subfigure}{\textwidth} 
        \centering
        \includegraphics[width=0.23\textwidth]{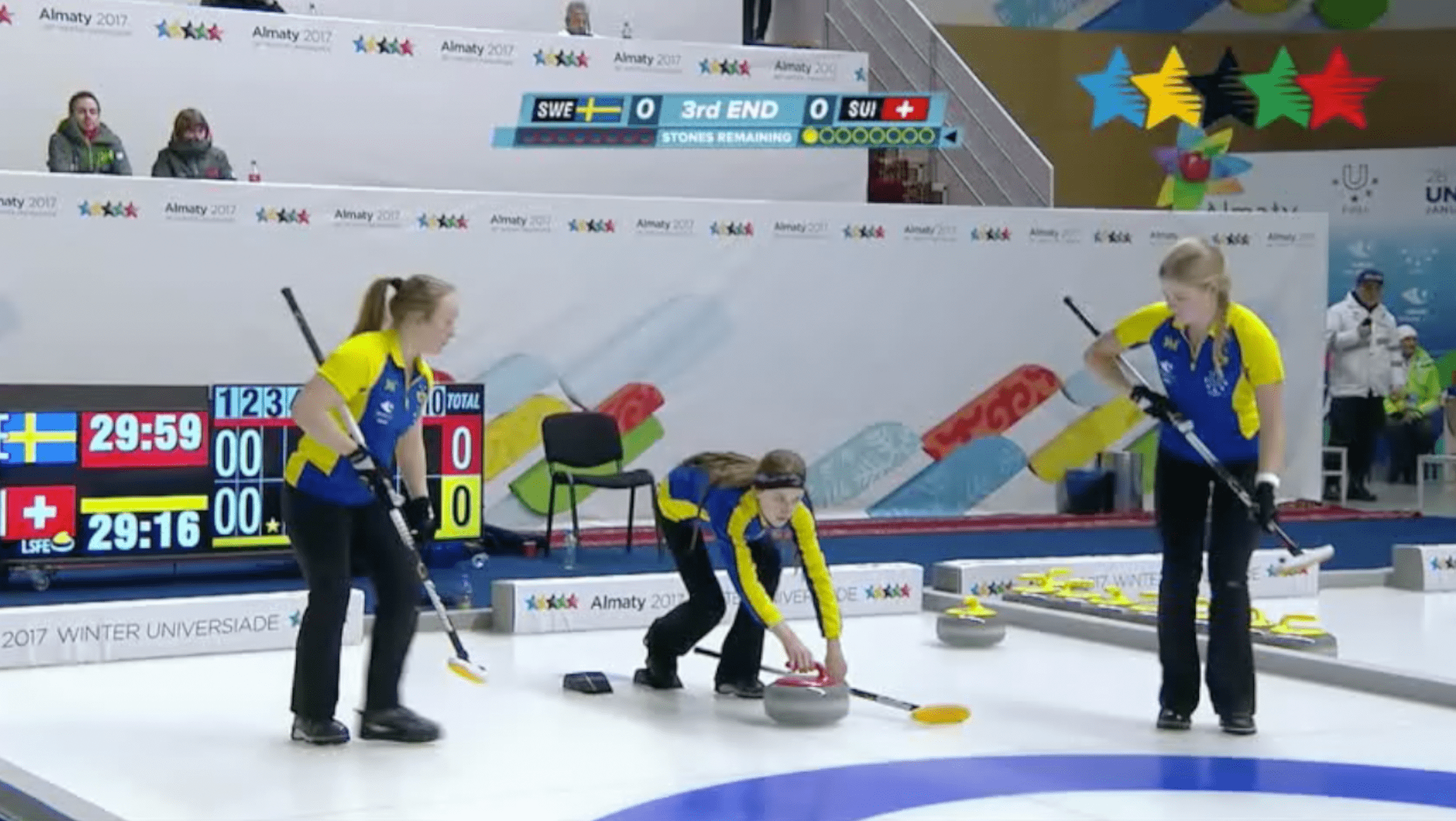}
        \includegraphics[width=0.23\textwidth]{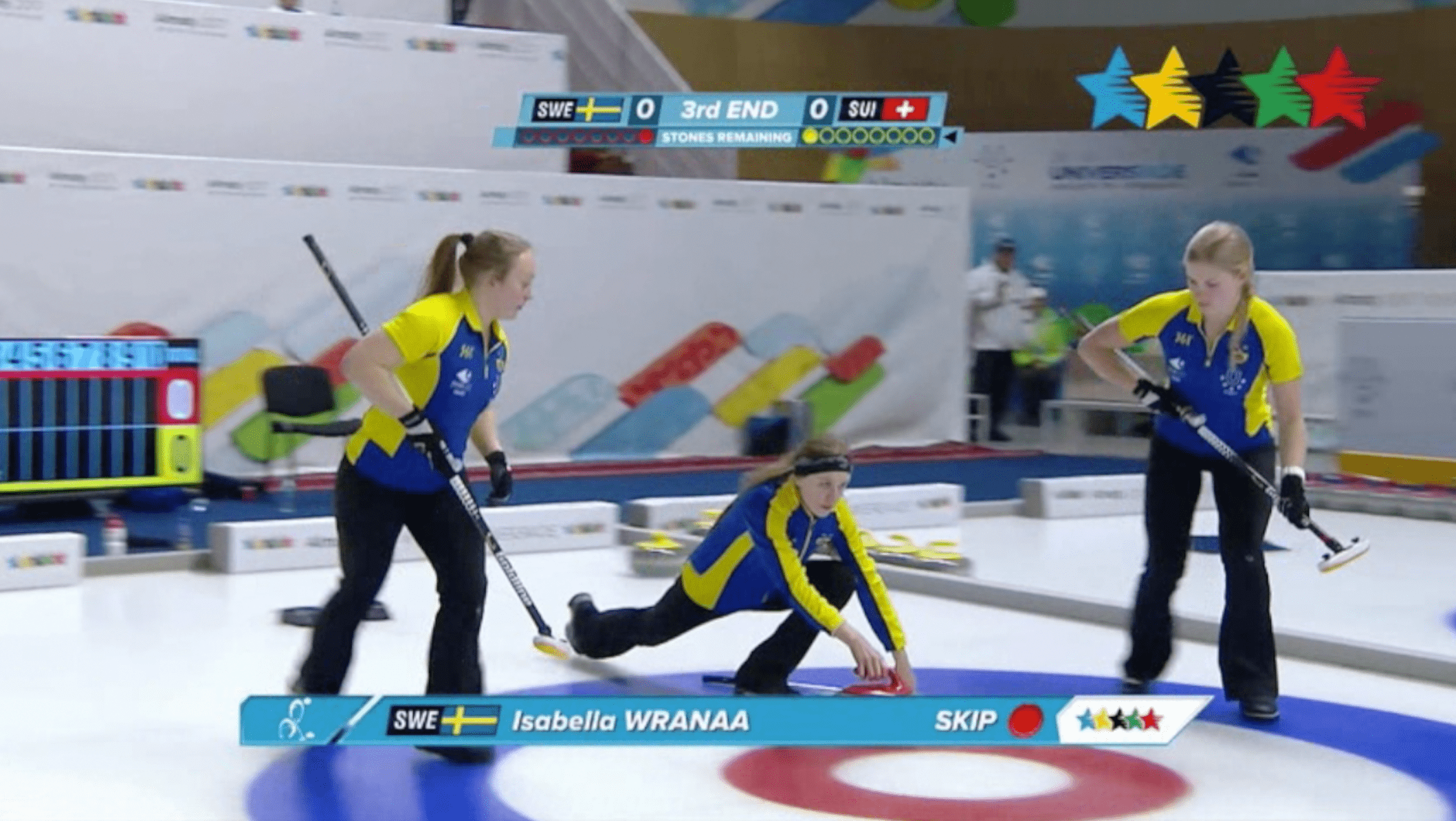}
        \includegraphics[width=0.23\textwidth]{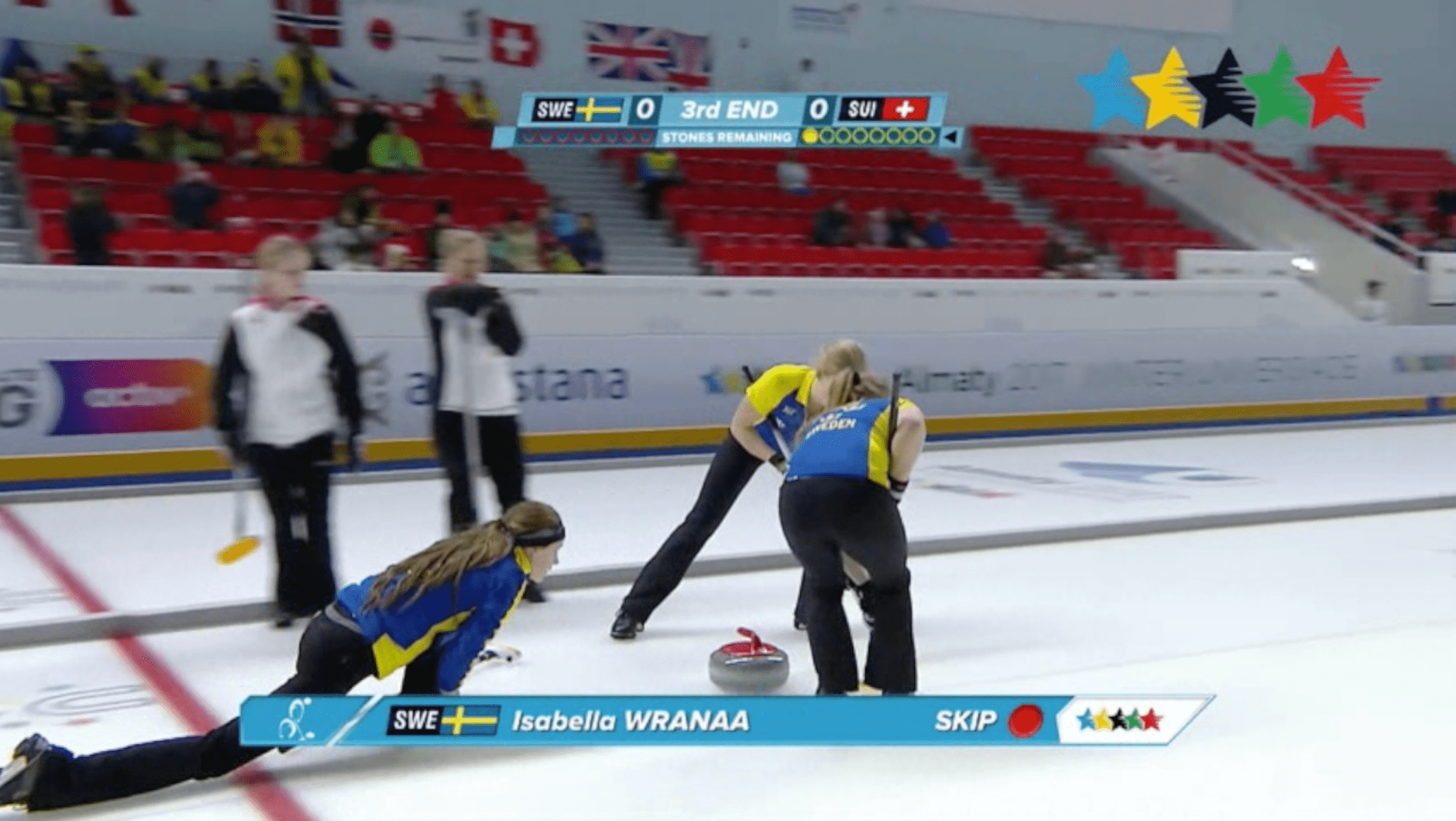}
        \includegraphics[width=0.23\textwidth]{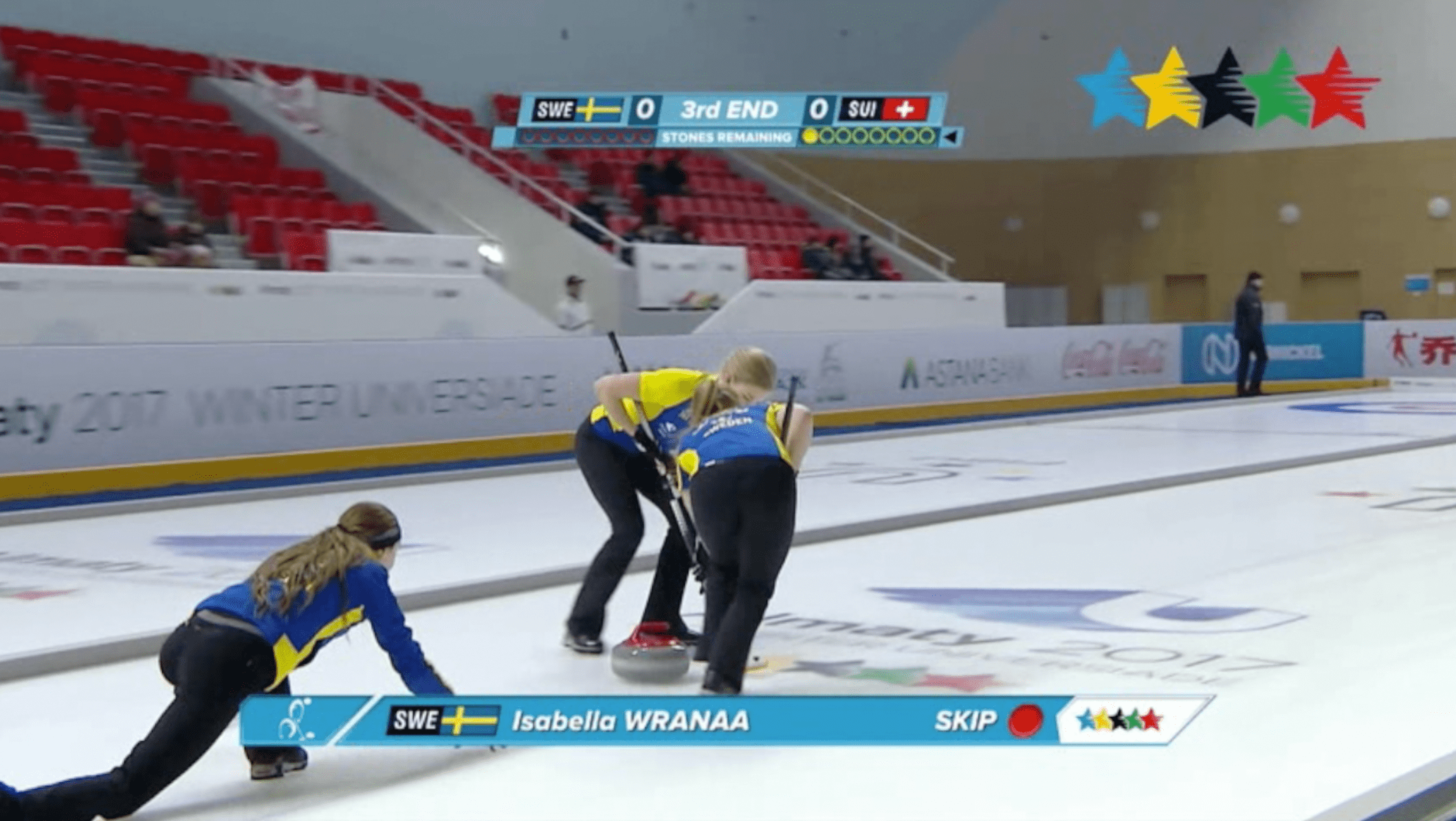}
        \caption{Curling (Exo-centric)}
    \end{subfigure}%
    \vspace{0.5em}
    \begin{subfigure}{\textwidth}
        \centering
        \includegraphics[width=0.23\textwidth]{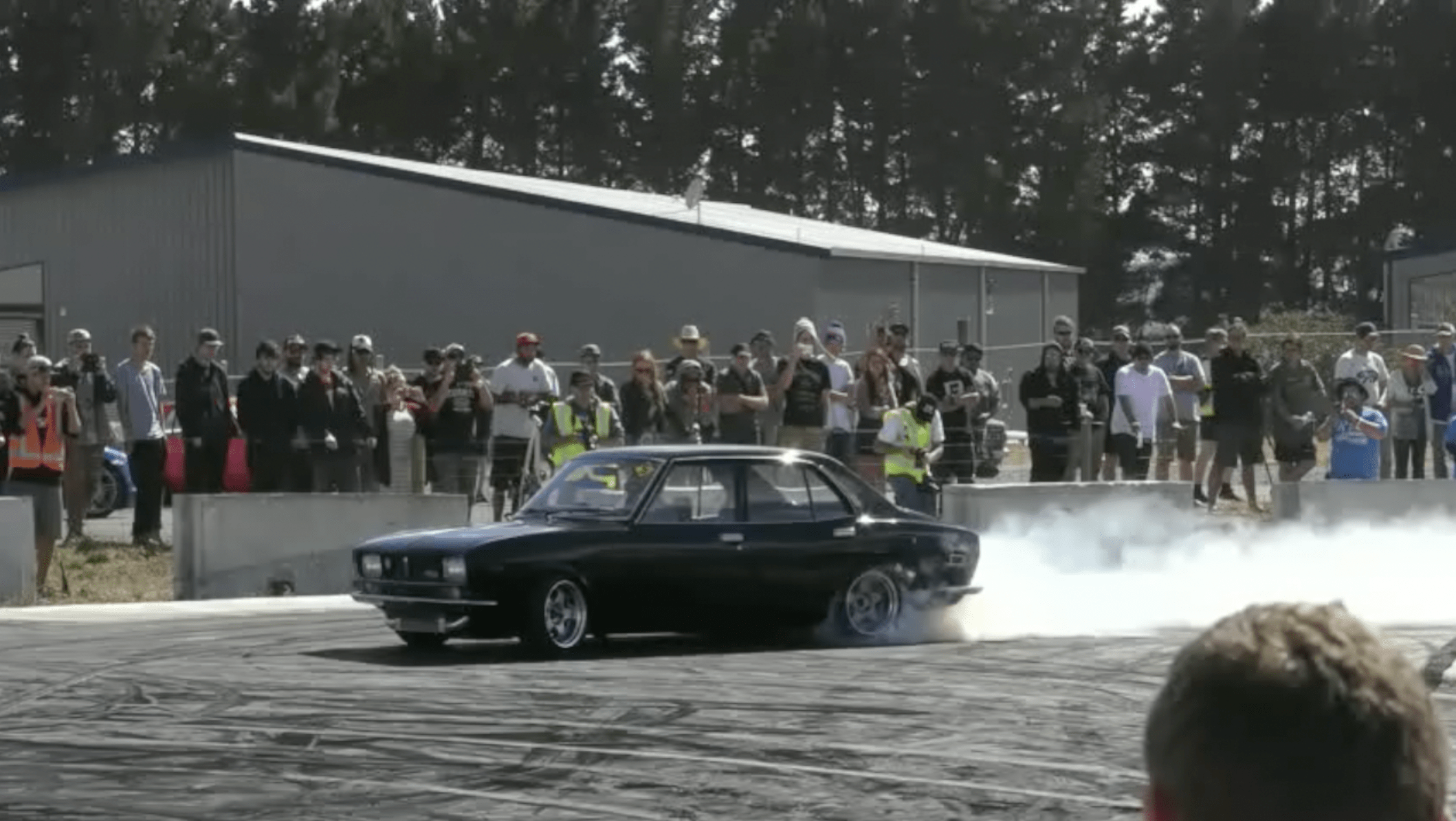}
        \includegraphics[width=0.23\textwidth]{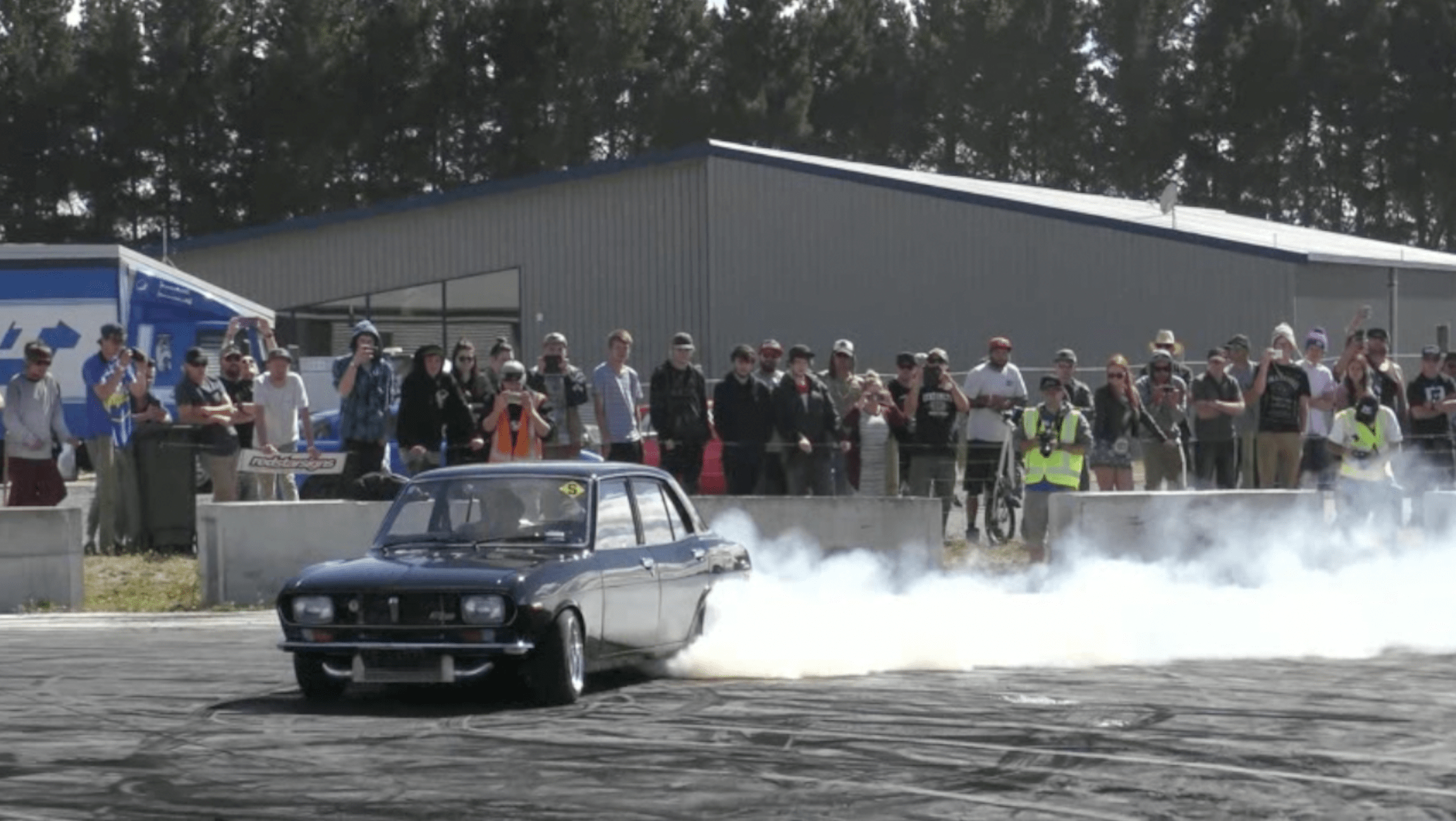}
        \includegraphics[width=0.23\textwidth]{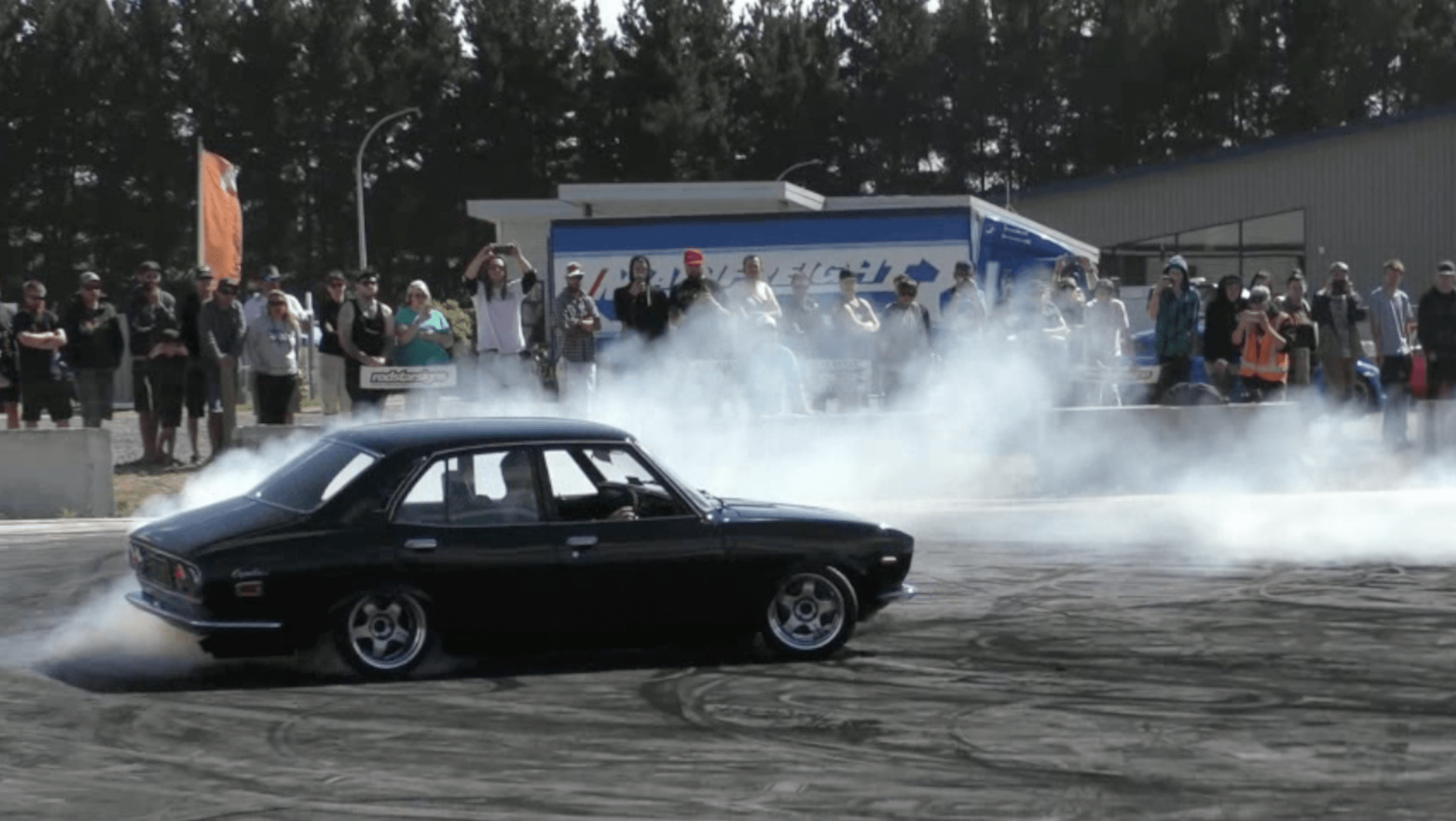}
        \includegraphics[width=0.23\textwidth]{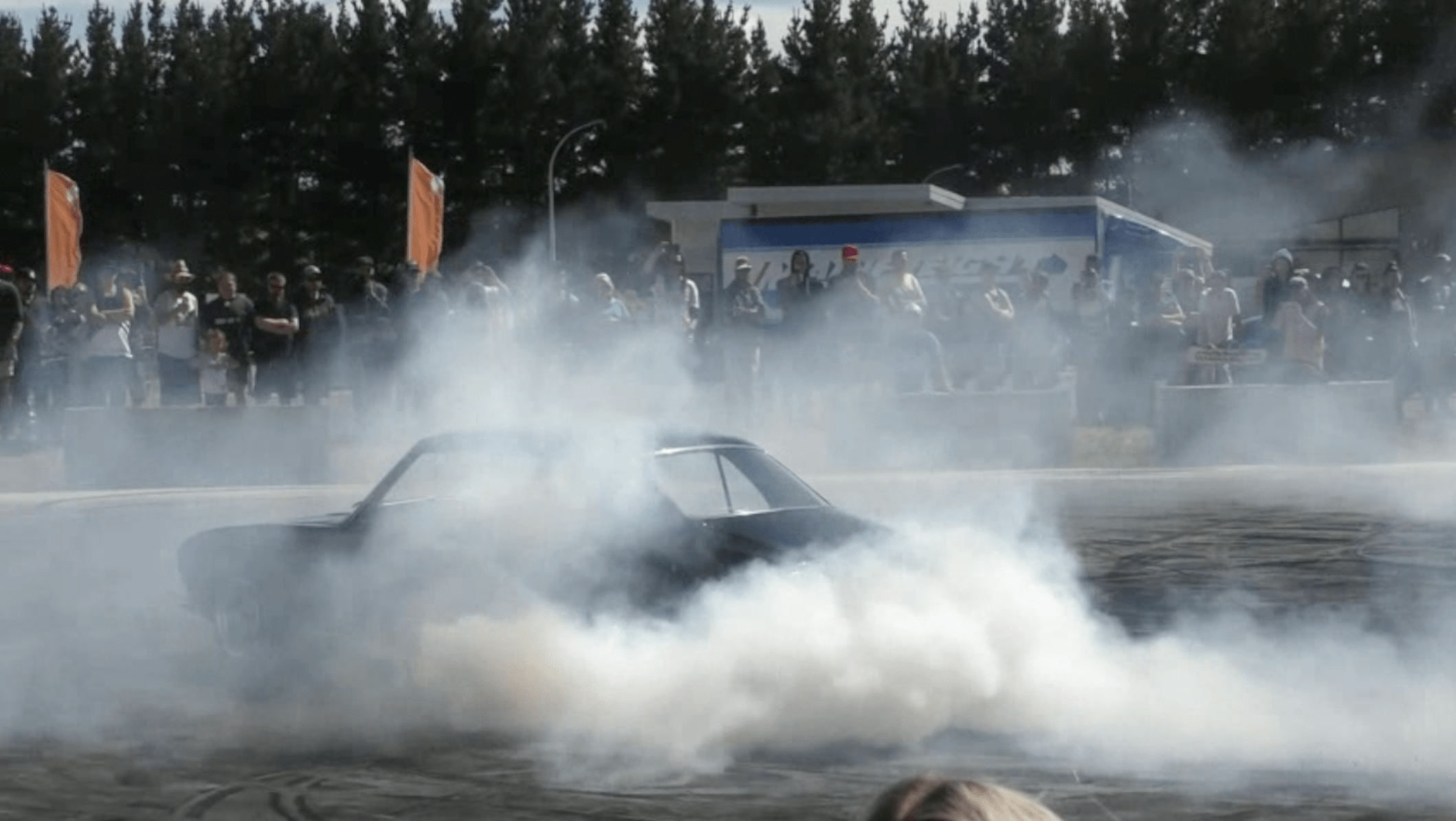}
        \caption{Burnout (Exo-centric)}
    \end{subfigure}%
    \vspace{0.5em}
    \begin{subfigure}{\textwidth}
        \centering
        \includegraphics[width=0.23\textwidth]{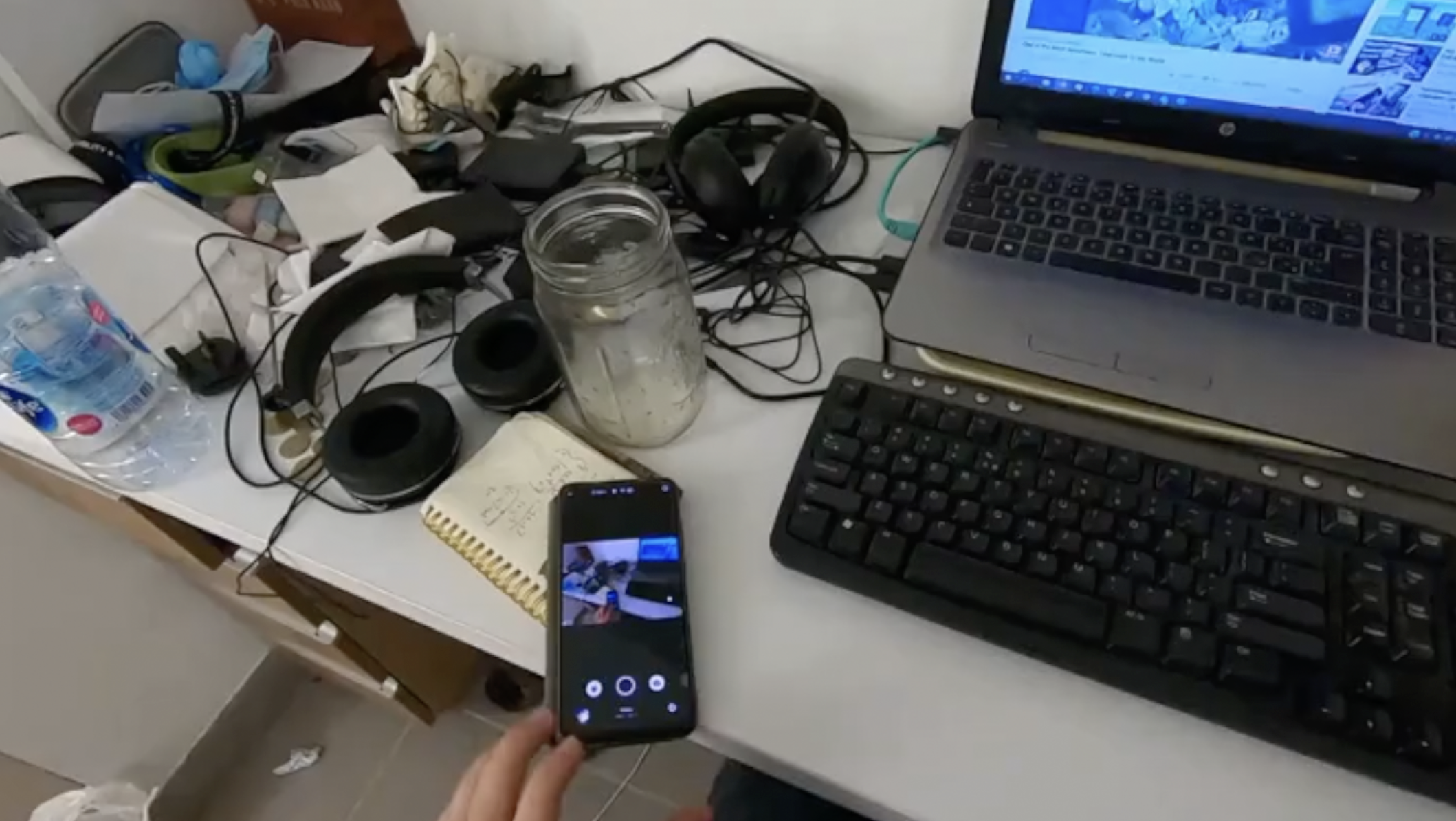}
        \includegraphics[width=0.23\textwidth]{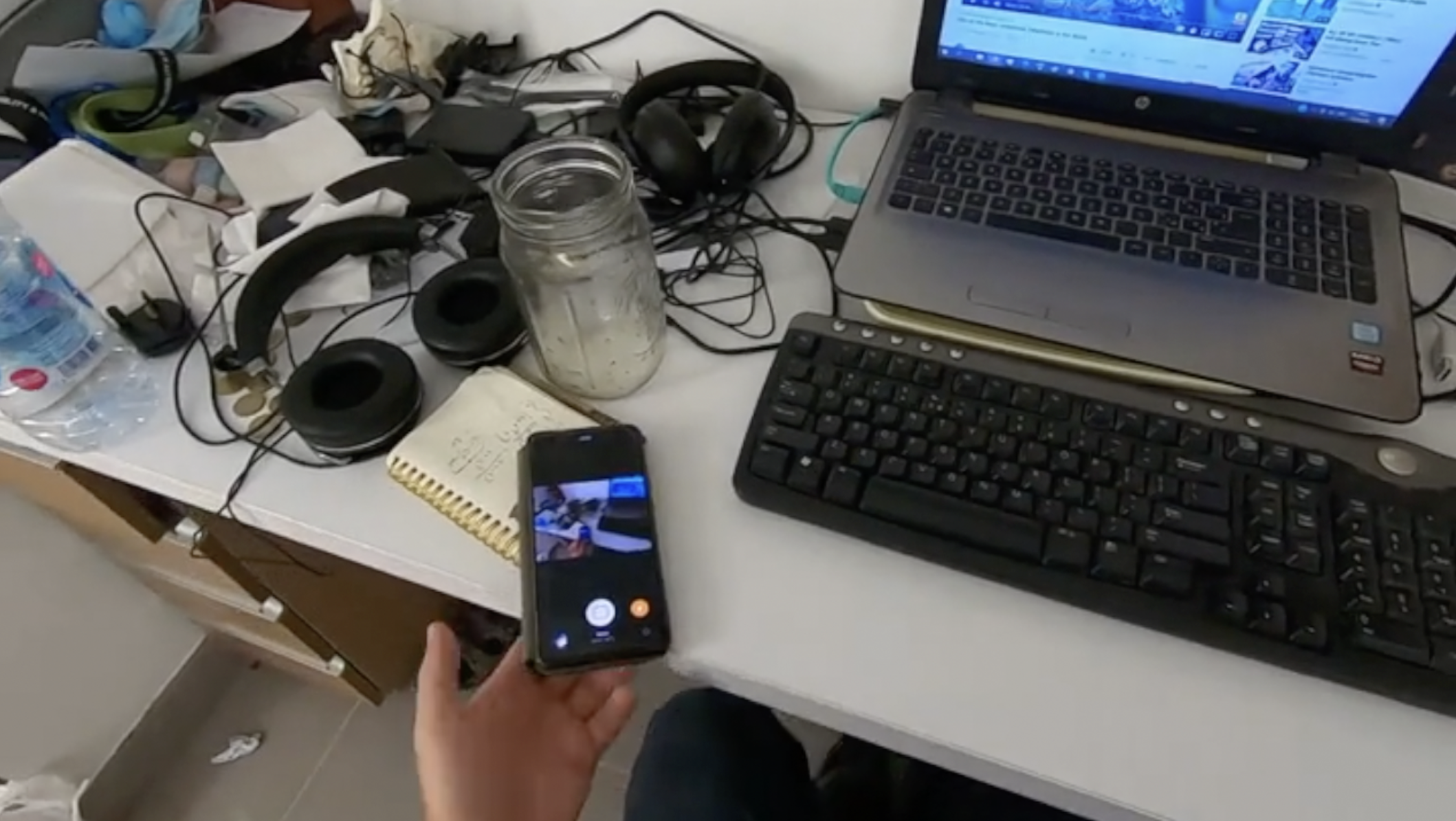}
        \includegraphics[width=0.23\textwidth]{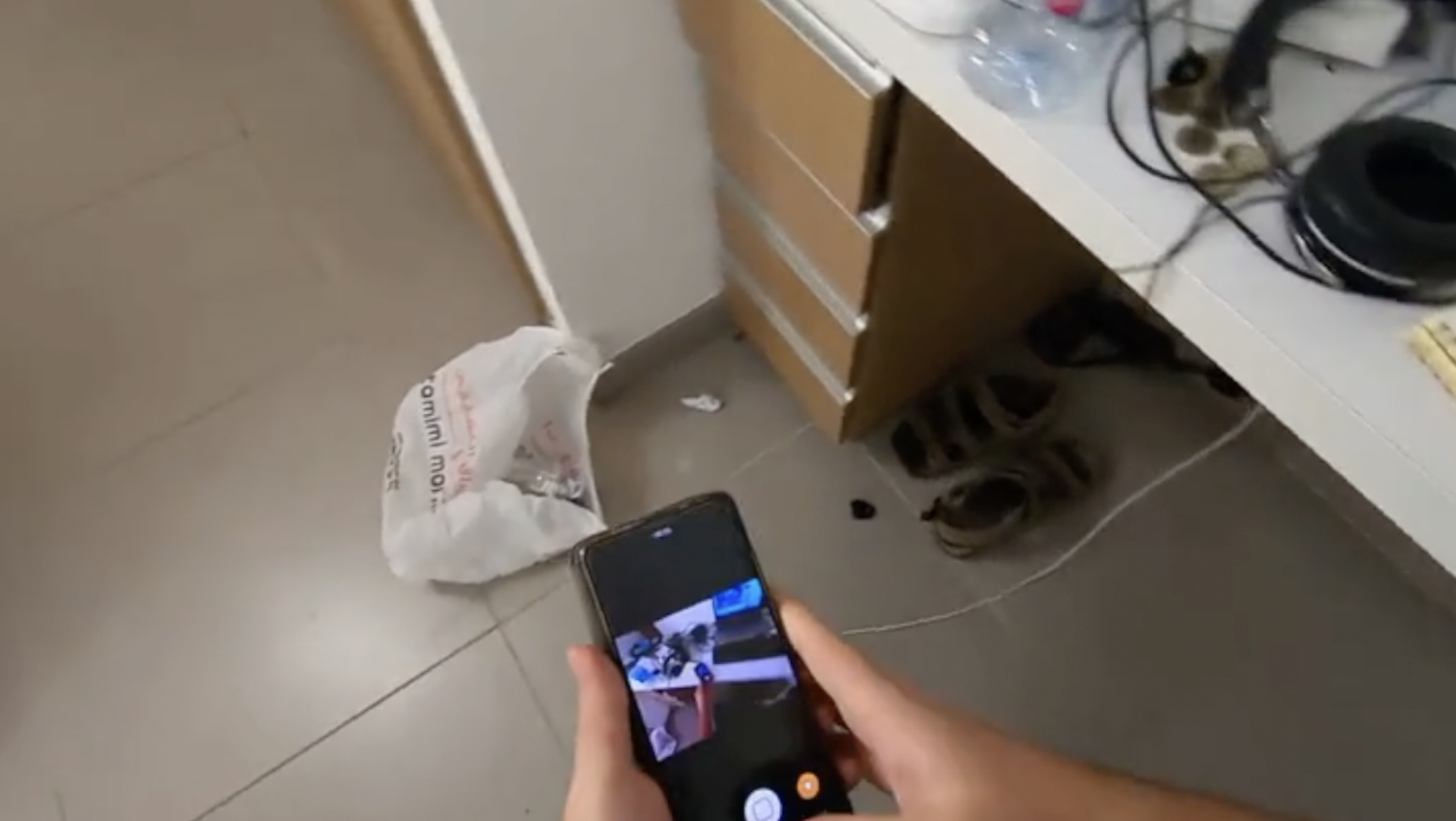}
        \includegraphics[width=0.23\textwidth]{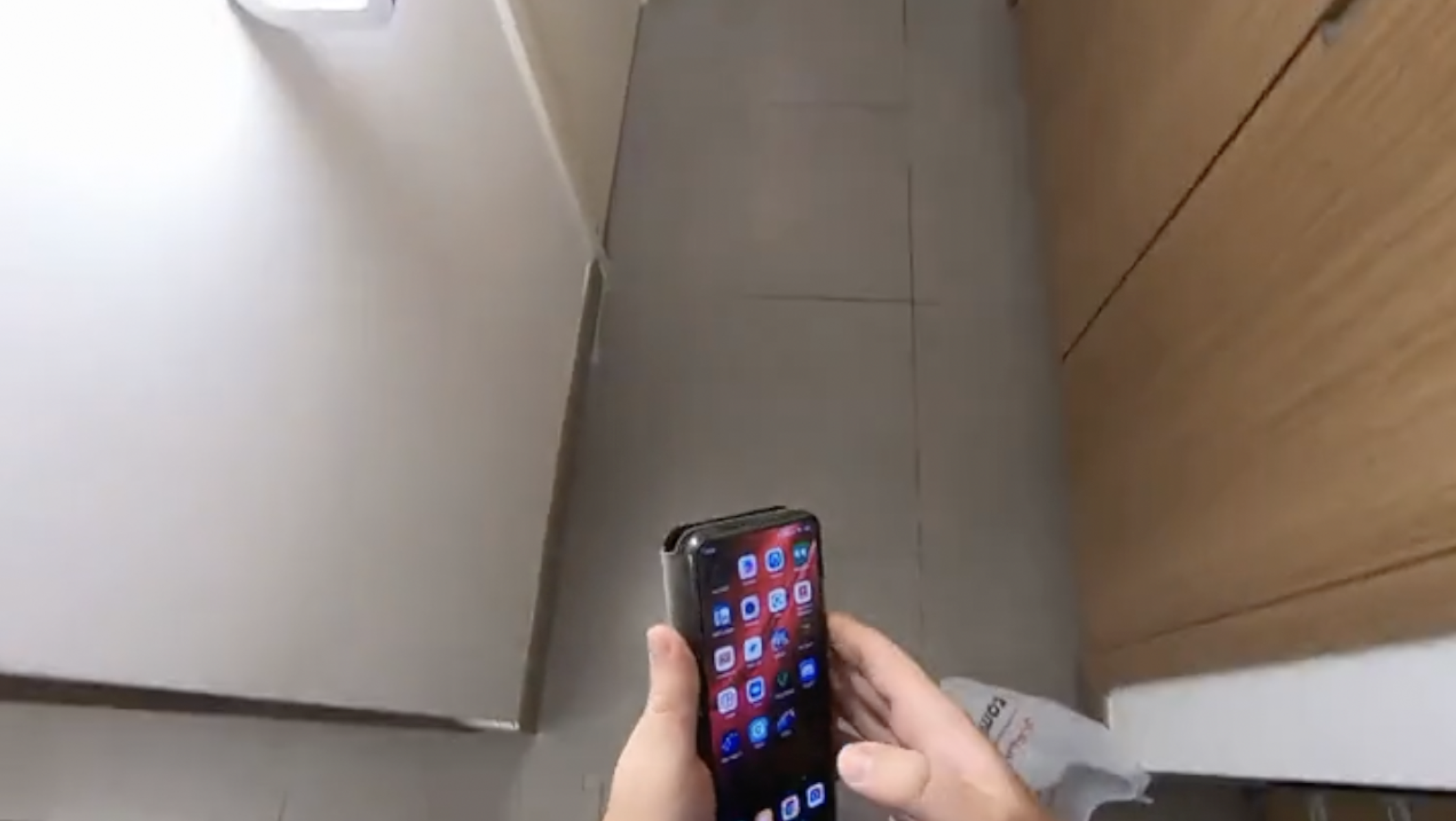}
        \caption{Desk (Ego-centric)}
    \end{subfigure}%
    \vspace{0.5em}
    \begin{subfigure}{\textwidth}
        \centering
        \includegraphics[width=0.23\textwidth]{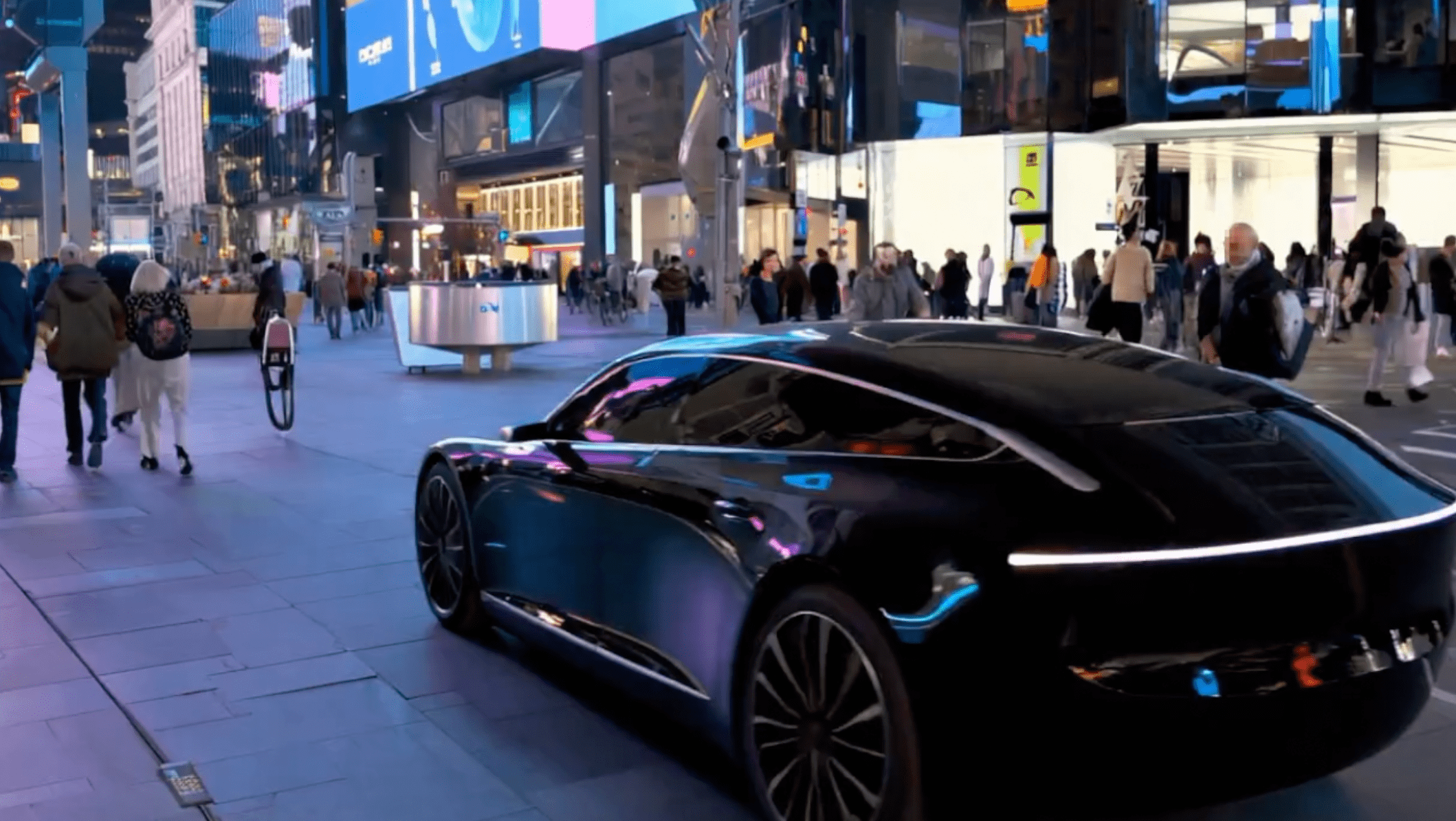}
        \includegraphics[width=0.23\textwidth]{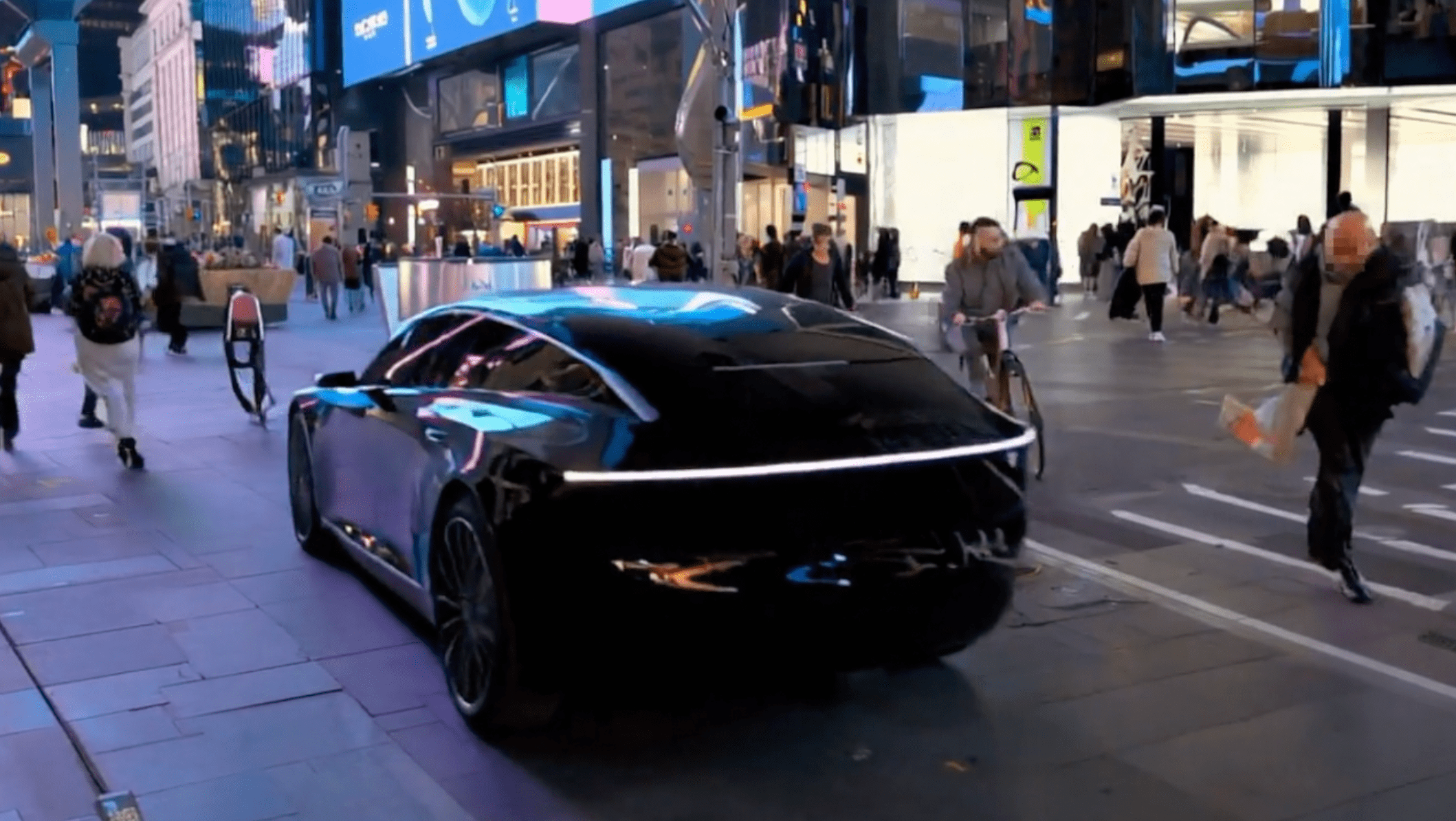}
        \includegraphics[width=0.23\textwidth]{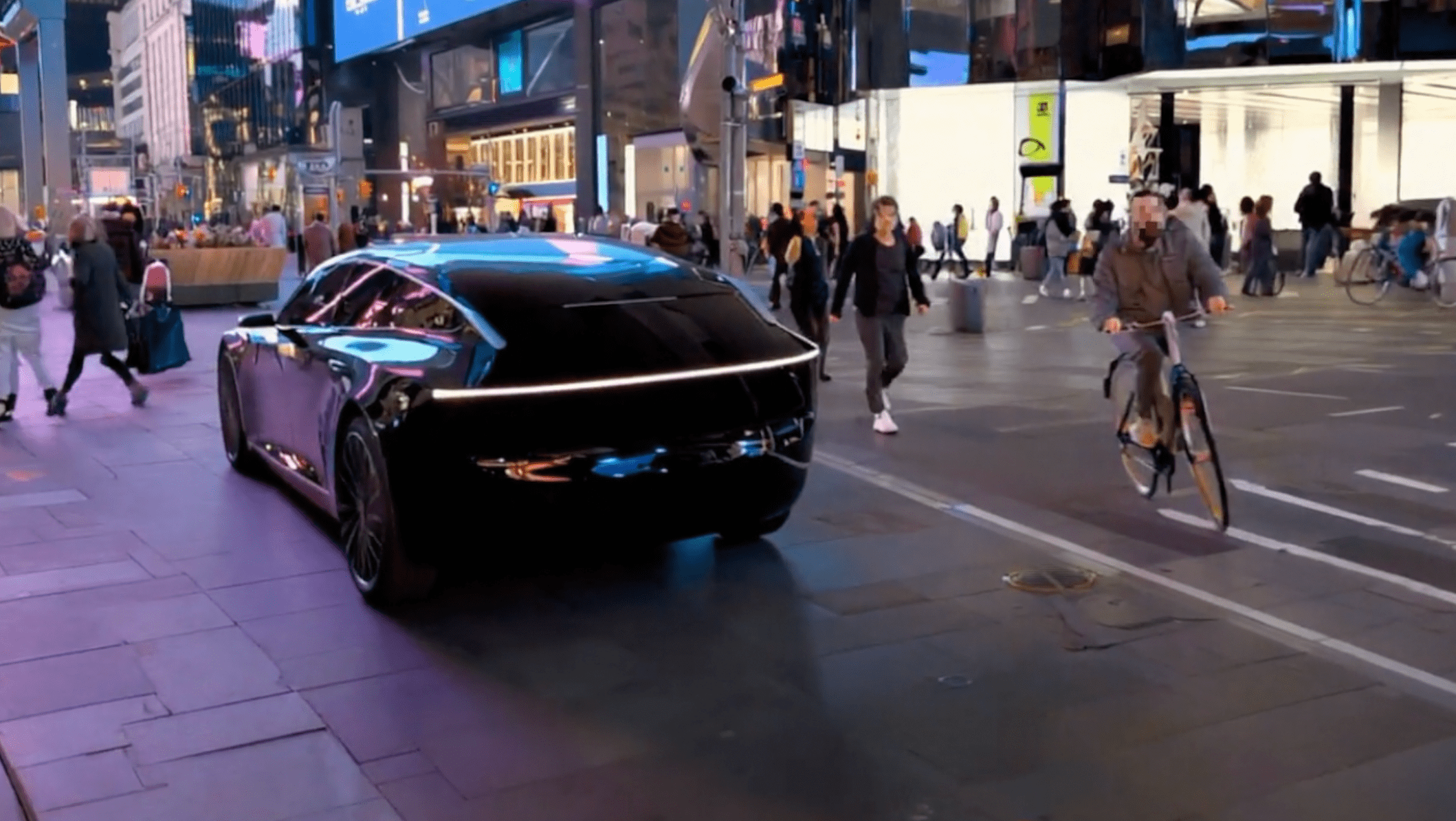}
        \includegraphics[width=0.23\textwidth]{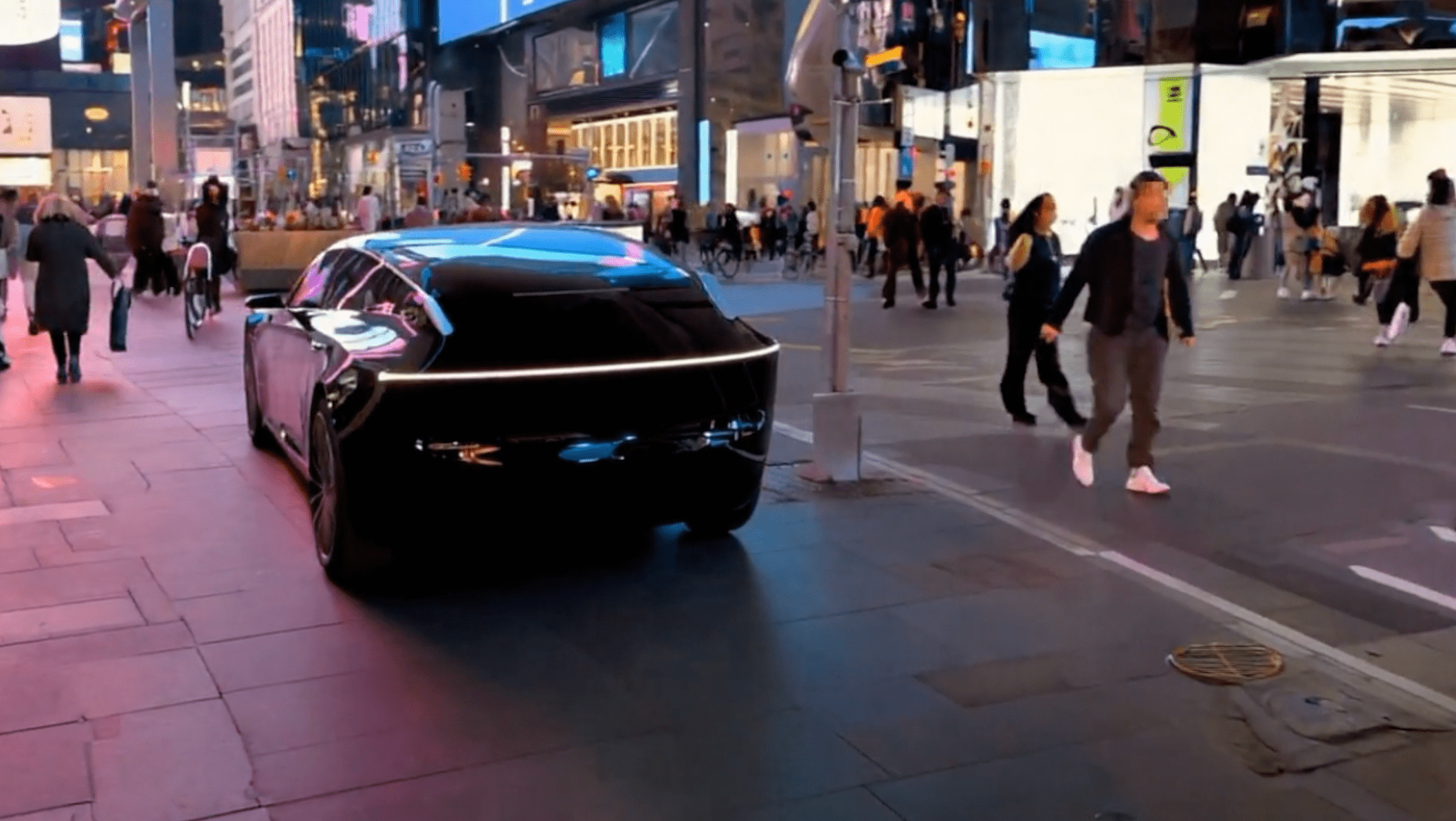}
        \caption{Futuristic Car(Synthetic)}
    \end{subfigure}%
    \caption{\textbf{Qualitative examples of SNOW on the VLM4D benchmark} across exo-centric, ego-centric, and synthetic videos. Each example displays samples of one video trace alongside its question, ground truth and predicted answer provided in Table~\ref{tab:vlm4d_multiframe_examples} to illustrate correct and failure cases of SNOW.}
    \label{fig:vlm4d_multiframe_examples}
\end{figure*}

\begin{table*}[!t]
    \small
    \centering
    \begin{tabular}{c|c|p{0.40\textwidth}|c|c}
        \toprule
        \textbf{Scenario} & \textbf{ID} & \textbf{VLM4D Question} & \textbf{Ground Truth} & \textbf{SNOW's Answer} \\
        \midrule
        \multirow{6}{*}{\textbf{Curling}} 
            & Q.182 & How many people are moving to the right on the ice rink? & 3 & 3 \\
            & Q.184 & From the camera perspective, which direction is the curling team moving towards? & right & right \\
            & Q.186 & How many people are sweeping in front of the moving curling stone? & 2 & 2 \\
        \midrule
        \multirow{7}{*}{\textbf{Burnout}} 
            & Q.77 & Is the car spinning clockwise or counter-clockwise? & counter-clockwise & clockwise \\
            & Q.78 & From the camera perspective, what direction is the car moving towards? & left & right \\
            & Q.79 & From the cars perspective, is it turning to the left or right? & left & right \\
            & Q.80 & Which direction is the crowd in the background moving towards? & not moving & not moving \\
        \midrule
        \multirow{3}{*}{\textbf{Desk}} 
            & Q.1242 & What does the left hand do? & pick up the phone & pick up the phone \\
            & Q.1243 & What does the right hand do? & hold the phone & hold the phone \\
            & Q.1244 & What direction is the table moving? & not moving & not moving \\
        \midrule
        \multirow{2}{*}{\textbf{Futuristic Car}} 
            & Q.61 & What direction is the fairy moving towards? & no fairy there & no fairy there \\
            & Q.62 & What direction is the taxi moving towards? & left & no taxi there \\
        \bottomrule
    \end{tabular}
    \caption{\textbf{Question formulations on VLM4D} corresponding to video traces in Figure~\ref{fig:vlm4d_multiframe_examples}. For each scenario we select representative questions to qualitatively illustrate the answers of SNOW.}
    \label{tab:vlm4d_multiframe_examples}
\end{table*}

\section{Qualitative Examples for open-vocabulary LiDAR Segmentation} 
\label{appendix:nuscens}

\begin{figure*}[!t]
    \centering
    \begin{subfigure}{\textwidth} 
        \centering
        \includegraphics[width=0.31\textwidth]{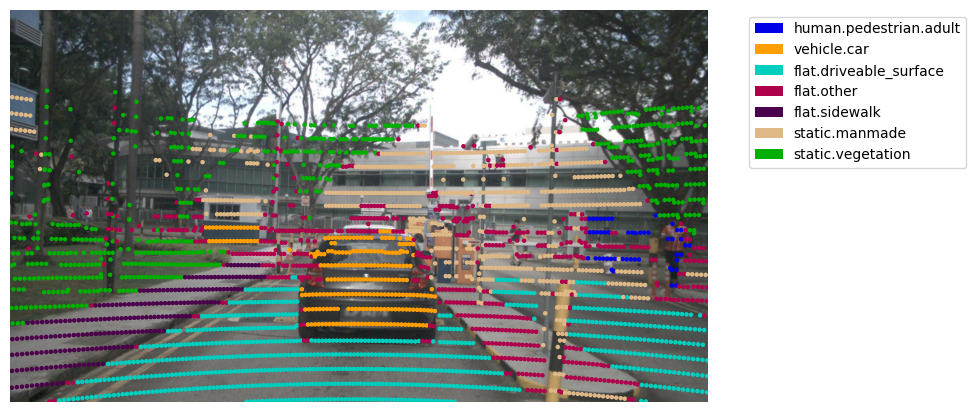}
        \includegraphics[width=0.31\textwidth]{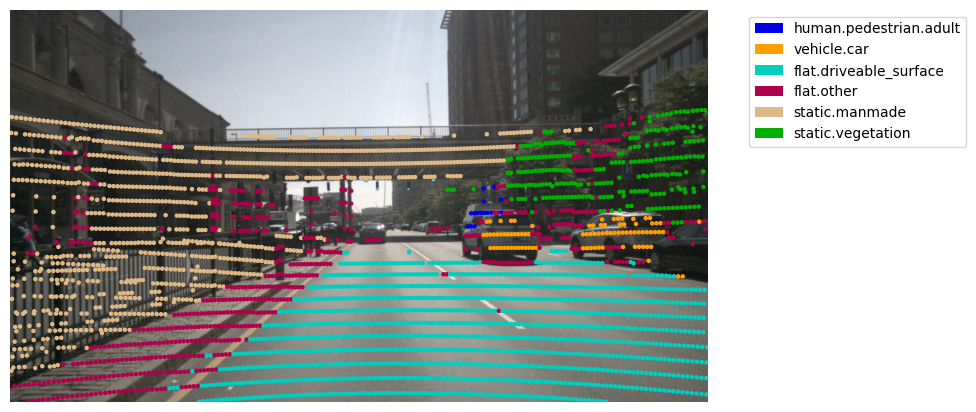}
        \includegraphics[width=0.31\textwidth]{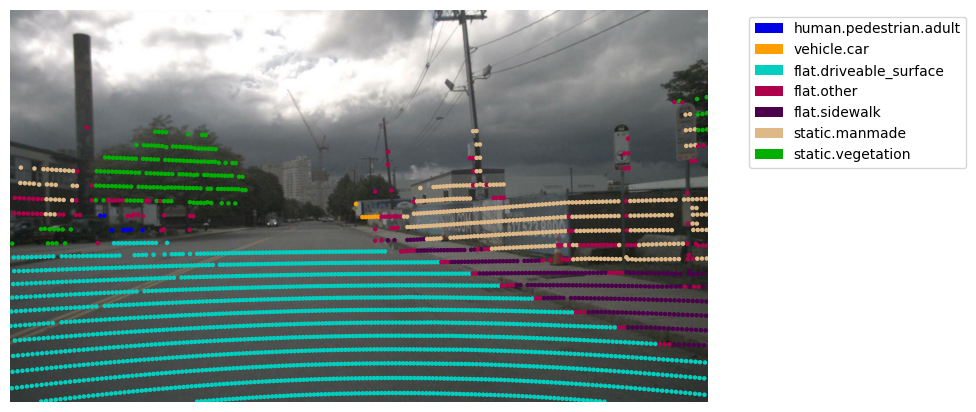}
    \end{subfigure}%
    \vspace{0.5em}
    \begin{subfigure}{\textwidth}
        \centering
        \includegraphics[width=0.31\textwidth]{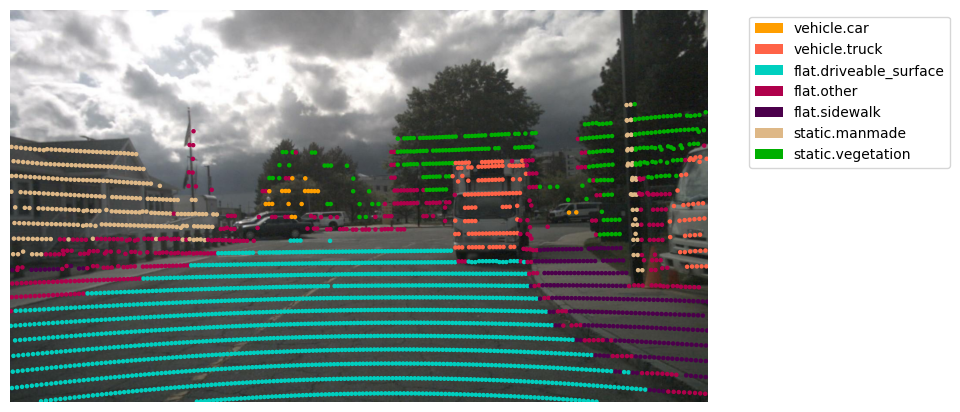}
        \includegraphics[width=0.31\textwidth]{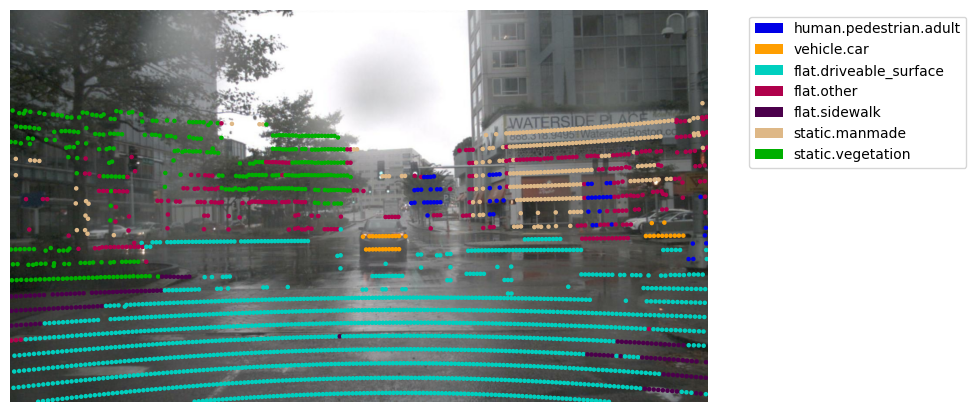}
        \includegraphics[width=0.31\textwidth]{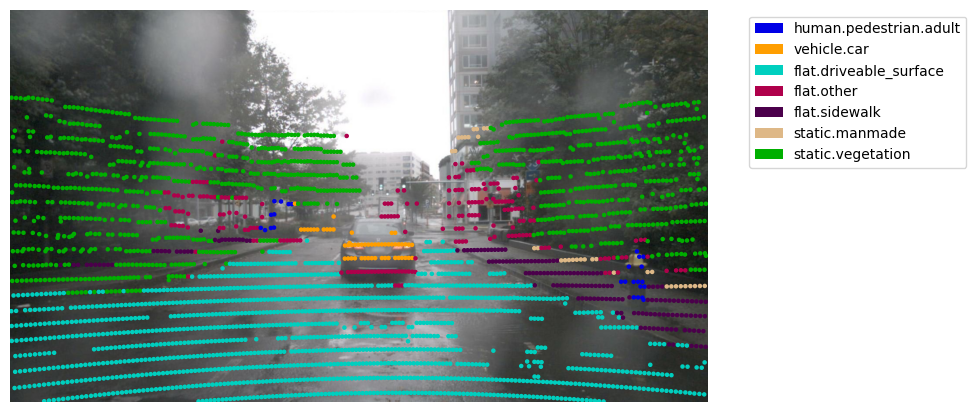}
    \end{subfigure}%
    \vspace{0.5em}
    \begin{subfigure}{\textwidth}
        \centering
        \includegraphics[width=0.31\textwidth]{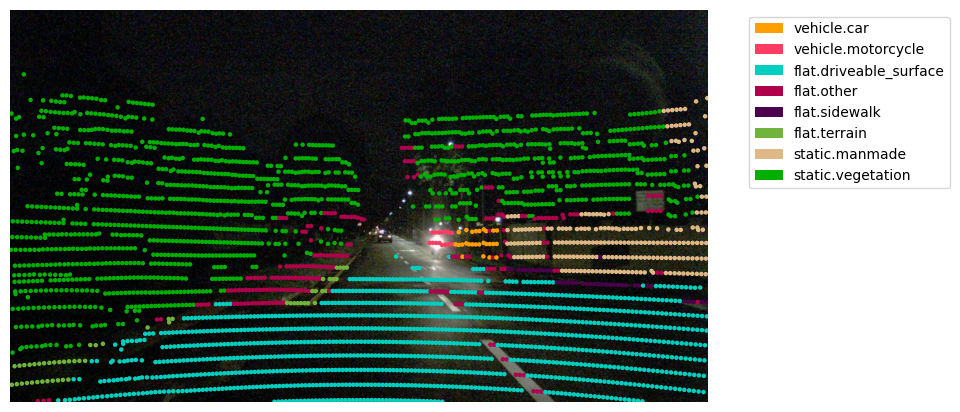}
        \includegraphics[width=0.31\textwidth]{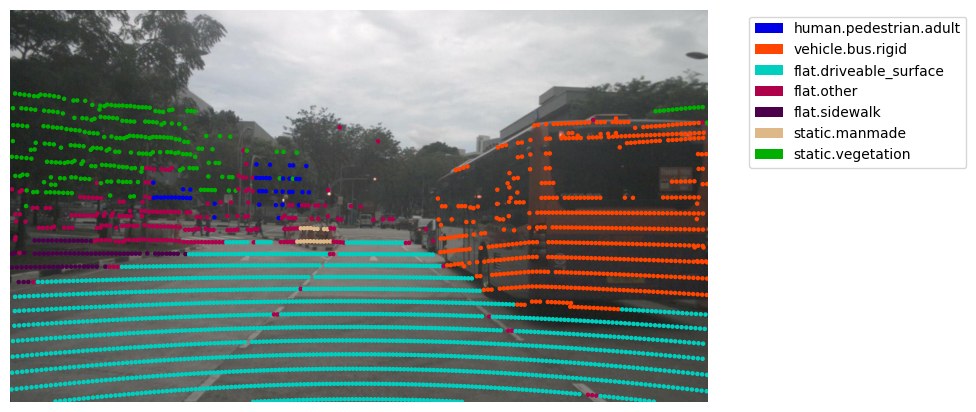}
        \includegraphics[width=0.31\textwidth]{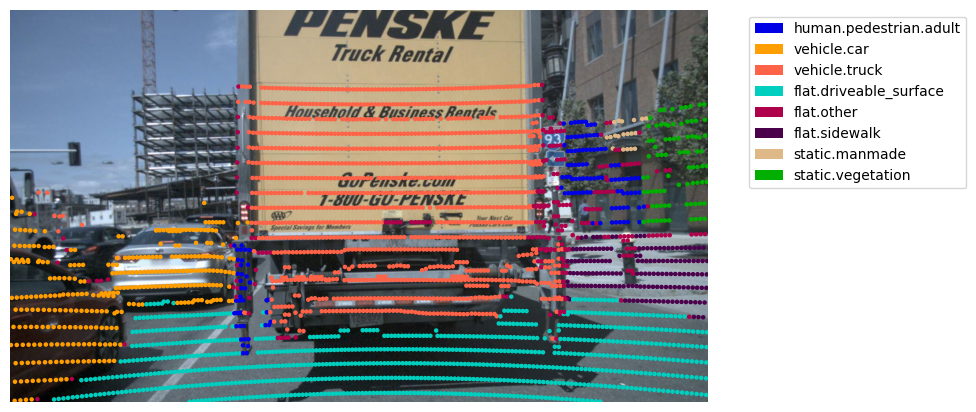}
    \end{subfigure}%
    \vspace{0.5em}
    \begin{subfigure}{\textwidth}
        \centering
        \includegraphics[width=0.31\textwidth]{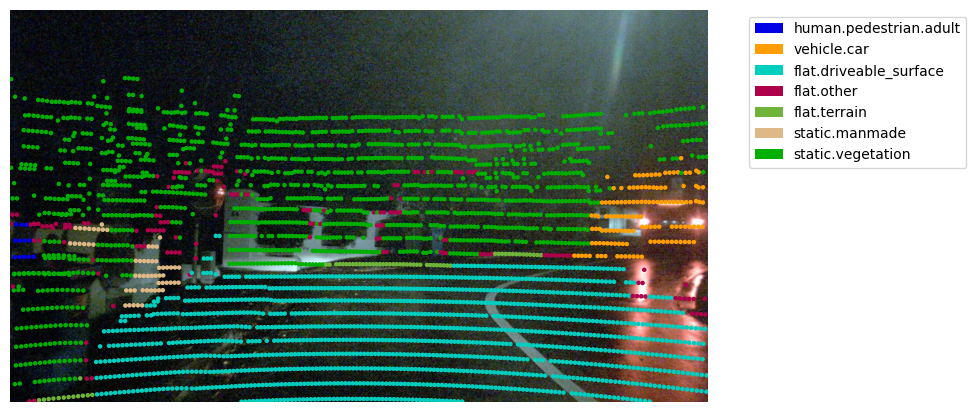}
        \includegraphics[width=0.31\textwidth]{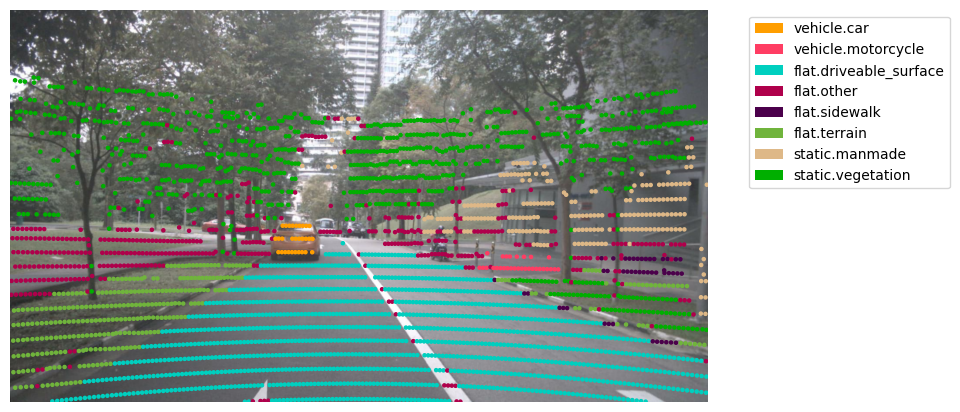}
        \includegraphics[width=0.31\textwidth]{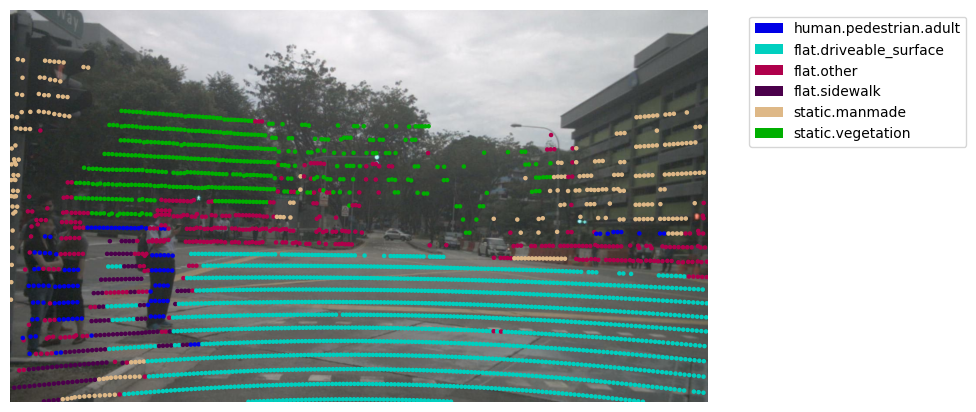}
    \end{subfigure}%
    \caption{\textbf{Qualitative examples of SNOW on the NuScenes LiDAR segmentation task} illustrating examples across diverse scenes, weather, and daylight conditions.}
    \label{fig:nuscenes_appendix}
\end{figure*}

Figure~\ref{fig:nuscenes_appendix} presents qualitative results of SNOW on the NuScenes LiDAR segmentation task~\cite{Panoptic_nuScenes_2021_Lidarseg}. SNOW segments single objects accurately by leveraging the spatially grounded 4D STEP tokens, which provide consistent object identities and geometry across frames without any task-specific training. This illustrates that the STEP representation alone is sufficient to transfer semantic associations from the world knowledge of VLMs in the image domain into the LiDAR space.

Smaller errors typically occur at fine object boundaries or in cluttered scenes with small, partially occluded instances. Since SNOW does not learn class-specific point-level features, it is less effective when geometric cues are weak or objects lack distinct volumetric separation for semantic segmentation. Nonetheless, the overall qualitative behavior confirms that structured 4D spatial grounding enables meaningful 3D segmentation performance even in a training-free setting, demonstrating the versatility and generality of the STEP representation beyond 4D language reasoning.

\section{Runtime Considerations} 
\label{appendix:runtime}
\begin{figure}
    \centering
    \includegraphics[width=\linewidth]{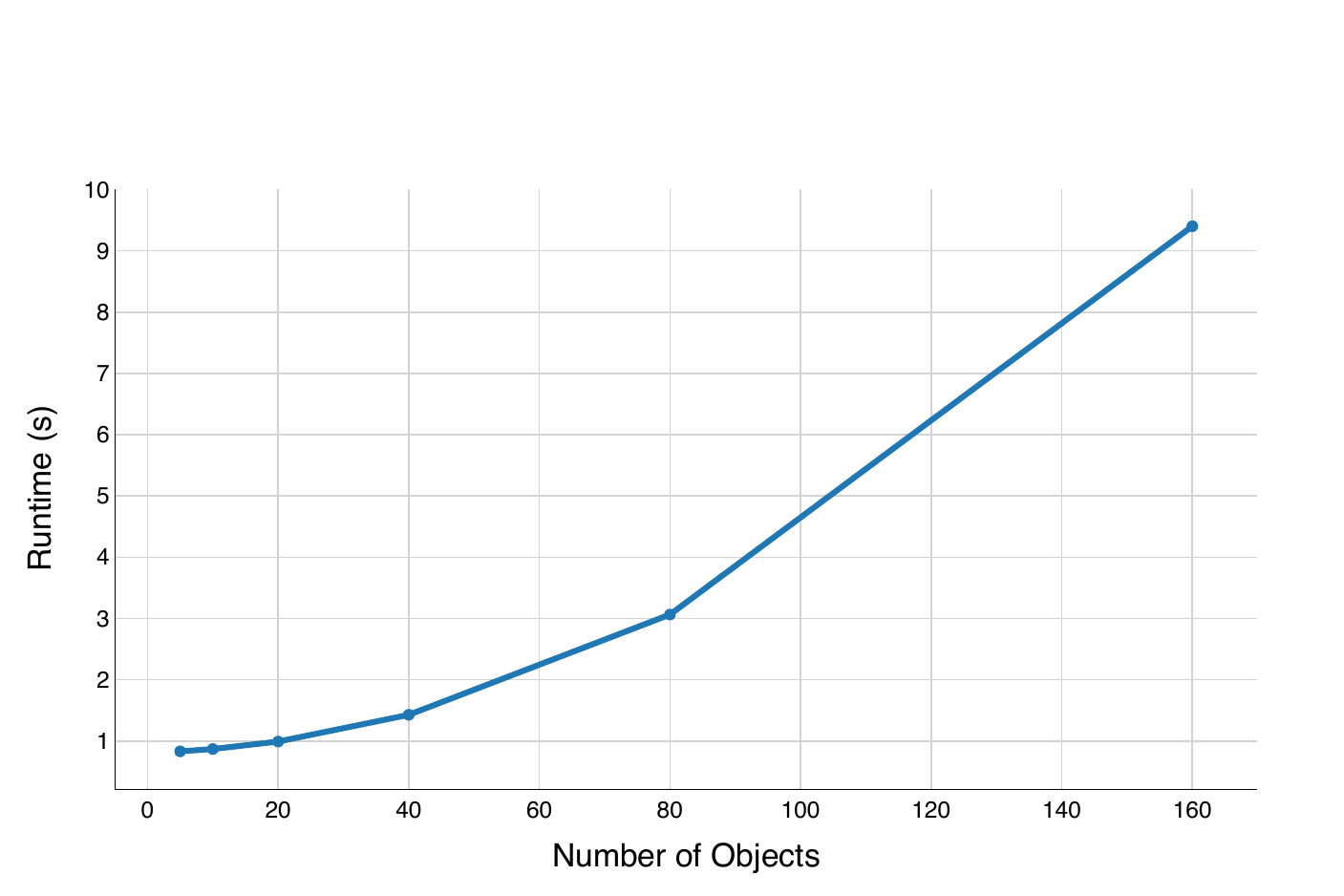}
    \caption{\textbf{Runtime scaling of SNOW} as a function of the number of integrated segmentation masks. The curve illustrates how increasing mask density impacts computational cost under identical inference settings.}
    \label{fig:runtime}
\end{figure}

Runtime is evaluated per frame on a single NVIDIA H100 GPU using batched inference across MapAnything~\cite{Keetha_2025_arXiv_MapAnything}, SAM2\_Hiera\_Large~\cite{Ravi_2024_arXiv_SAM2}, and Gemma3-4B-IT~\cite{Google_2025_Gemma3}. Figure~\ref{fig:runtime} reports the end-to-end processing time as a function of the number of segmented objects. The dominant overhead arises from the VLM's input context: as object count increases, the resulting 4DSGs grow and the per-frame latency rises accordingly. While the current implementation does not meet real-time requirements, the runtime remains practical for short-horizon embodied tasks that rely on 4D contextual reasoning and tactical scene understanding.

%% file: main.bib
@misc{Zhou_2025_arXiv_VLM4D,
      title={VLM4D: Towards Spatiotemporal Awareness in Vision Language Models}, 
      author={Shijie Zhou and Alexander Vilesov and Xuehai He and Ziyu Wan and Shuwang Zhang and Aditya Nagachandra and Di Chang and Dongdong Chen and Xin Eric Wang and Achuta Kadambi},
      year={2025},
      eprint={2508.02095},
      archivePrefix={arXiv},
      primaryClass={cs.CV},
      url={https://arxiv.org/abs/2508.02095}, 
}

@article{Fong_2021_Lidarseg,
  title={Panoptic nuScenes: A Large-Scale Benchmark for LiDAR Panoptic Segmentation and Tracking},
  author={Fong, Whye Kit and Mohan, Rohit and Hurtado, Juana Valeria and Zhou, Lubing and Caesar, Holger and
          Beijbom, Oscar and Valada, Abhinav},
  journal={arXiv preprint arXiv:2109.03805},
  year={2021}
}

@misc{Chen_2023_arXiv_CNS,
      title={Towards Label-free Scene Understanding by Vision Foundation Models}, 
      author={Runnan Chen and Youquan Liu and Lingdong Kong and Nenglun Chen and Xinge Zhu and Yuexin Ma and Tongliang Liu and Wenping Wang},
      year={2023},
      eprint={2306.03899},
      archivePrefix={arXiv},
      primaryClass={cs.CV},
      url={https://arxiv.org/abs/2306.03899}, 
}

@misc{Peng_2023_arXiv_OpenScene,
      title={OpenScene: 3D Scene Understanding with Open Vocabularies}, 
      author={Songyou Peng and Kyle Genova and Chiyu "Max" Jiang and Andrea Tagliasacchi and Marc Pollefeys and Thomas Funkhouser},
      year={2023},
      eprint={2211.15654},
      archivePrefix={arXiv},
      primaryClass={cs.CV},
      url={https://arxiv.org/abs/2211.15654}, 
}

@misc{Zou_2024_arXiv_AdaCo,
      title={AdaCo: Overcoming Visual Foundation Model Noise in 3D Semantic Segmentation via Adaptive Label Correction}, 
      author={Pufan Zou and Shijia Zhao and Weijie Huang and Qiming Xia and Chenglu Wen and Wei Li and Cheng Wang},
      year={2024},
      eprint={2412.18255},
      archivePrefix={arXiv},
      primaryClass={cs.CV},
      url={https://arxiv.org/abs/2412.18255}, 
}

@InProceedings{Wei_2025_CVPR_3dAVS,
    author    = {Wei, Weijie and \"Ulger, Osman and Nejadasl, Fatemeh Karimi and Gevers, Theo and Oswald, Martin R.},
    title     = {3D-AVS: LiDAR-based 3D Auto-Vocabulary Segmentation},
    booktitle = {Proceedings of the IEEE/CVF Conference on Computer Vision and Pattern Recognition (CVPR)},
    month     = {June},
    year      = {2025},
    pages     = {8910-8920}
}

@InProceedings{Jiang_2024_CVPR_OV3D,
    author    = {Jiang, Li and Shi, Shaoshuai and Schiele, Bernt},
    title     = {Open-Vocabulary 3D Semantic Segmentation with Foundation Models},
    booktitle = {Proceedings of the IEEE/CVF Conference on Computer Vision and Pattern Recognition (CVPR)},
    month     = {June},
    year      = {2024},
    pages     = {21284-21294}
}

@misc{Luo_2025_arXiv_NaviMaster,
      title={NaviMaster: Learning a Unified Policy for GUI and Embodied Navigation Tasks}, 
      author={Zhihao Luo and Wentao Yan and Jingyu Gong and Min Wang and Zhizhong Zhang and Xuhong Wang and Yuan Xie and Xin Tan},
      year={2025},
      eprint={2508.02046},
      archivePrefix={arXiv},
      primaryClass={cs.RO},
      url={https://arxiv.org/abs/2508.02046}, 
}

@misc{Song_2025_arXiv_RoboSpatial,
      title={RoboSpatial: Teaching Spatial Understanding to 2D and 3D Vision-Language Models for Robotics}, 
      author={Chan Hee Song and Valts Blukis and Jonathan Tremblay and Stephen Tyree and Yu Su and Stan Birchfield},
      year={2025},
      eprint={2411.16537},
      archivePrefix={arXiv},
      primaryClass={cs.CV},
      url={https://arxiv.org/abs/2411.16537}, 
}

@misc{Keetha_2025_arXiv_MapAnything,
      title={MapAnything: Universal Feed-Forward Metric 3D Reconstruction}, 
      author={Nikhil Keetha and Norman Müller and Johannes Schönberger and Lorenzo Porzi and Yuchen Zhang and Tobias Fischer and Arno Knapitsch and Duncan Zauss and Ethan Weber and Nelson Antunes and Jonathon Luiten and Manuel Lopez-Antequera and Samuel Rota Bulò and Christian Richardt and Deva Ramanan and Sebastian Scherer and Peter Kontschieder},
      year={2025},
      eprint={2509.13414},
      archivePrefix={arXiv},
      primaryClass={cs.CV},
      url={https://arxiv.org/abs/2509.13414}, 
}

@article{Panoptic_nuScenes_2021_Lidarseg,
  title={Panoptic nuScenes: A Large-Scale Benchmark for LiDAR Panoptic Segmentation and Tracking},
  author={Fong, Whye Kit and Mohan, Rohit and Hurtado, Juana Valeria and Zhou, Lubing and Caesar, Holger and
          Beijbom, Oscar and Valada, Abhinav},
  journal={arXiv preprint arXiv:2109.03805},
  year={2021}
}

@misc{Wei_2024_arXiv_OccLLaMA,
      title={OccLLaMA: An Occupancy-Language-Action Generative World Model for Autonomous Driving}, 
      author={Julong Wei and Shanshuai Yuan and Pengfei Li and Qingda Hu and Zhongxue Gan and Wenchao Ding},
      year={2024},
      eprint={2409.03272},
      archivePrefix={arXiv},
      primaryClass={cs.CV},
      url={https://arxiv.org/abs/2409.03272}, 
}

@misc{Gao_2023_arXiv_LLaMA-AdapV2,
      title={LLaMA-Adapter V2: Parameter-Efficient Visual Instruction Model}, 
      author={Peng Gao and Jiaming Han and Renrui Zhang and Ziyi Lin and Shijie Geng and Aojun Zhou and Wei Zhang and Pan Lu and Conghui He and Xiangyu Yue and Hongsheng Li and Yu Qiao},
      year={2023},
      eprint={2304.15010},
      archivePrefix={arXiv},
      primaryClass={cs.CV},
      url={https://arxiv.org/abs/2304.15010}, 
}

@InProceedings{Liu_2024_CVPR_LLaVA1.5,
    author    = {Liu, Haotian and Li, Chunyuan and Li, Yuheng and Lee, Yong Jae},
    title     = {Improved Baselines with Visual Instruction Tuning},
    booktitle = {Proceedings of the IEEE/CVF Conference on Computer Vision and Pattern Recognition (CVPR)},
    month     = {June},
    year      = {2024},
    pages     = {26296-26306}
}

@misc{Zhou_2025_arXiv_OpenDriveVLA,
      title={OpenDriveVLA: Towards End-to-end Autonomous Driving with Large Vision Language Action Model}, 
      author={Xingcheng Zhou and Xuyuan Han and Feng Yang and Yunpu Ma and Alois C. Knoll},
      year={2025},
      eprint={2503.23463},
      archivePrefix={arXiv},
      primaryClass={cs.CV},
      url={https://arxiv.org/abs/2503.23463}, 
}

@misc{Li_2024_arXiv_VideoChat_Temporal,
      title={VideoChat: Chat-Centric Video Understanding}, 
      author={KunChang Li and Yinan He and Yi Wang and Yizhuo Li and Wenhai Wang and Ping Luo and Yali Wang and Limin Wang and Yu Qiao},
      year={2024},
      eprint={2305.06355},
      archivePrefix={arXiv},
      primaryClass={cs.CV},
      url={https://arxiv.org/abs/2305.06355}, 
}

@misc{Lin_2024_arXiv_VideoLLaVA_Temporal,
      title={Video-LLaVA: Learning United Visual Representation by Alignment Before Projection}, 
      author={Bin Lin and Yang Ye and Bin Zhu and Jiaxi Cui and Munan Ning and Peng Jin and Li Yuan},
      year={2024},
      eprint={2311.10122},
      archivePrefix={arXiv},
      primaryClass={cs.CV},
      url={https://arxiv.org/abs/2311.10122}, 
}

@InProceedings{Liu_2024_ECCV_ST-LLM_Temporal,
    author="Liu, Ruyang and Li, Chen and Tang, Haoran and Ge, Yixiao and Shan, Ying and Li, Ge",
    editor="Leonardis, Ale{\v{s}} and Ricci, Elisa and Roth, Stefan and Russakovsky, Olga and Sattler, Torsten and Varol, G{\"u}l",
    title="ST-LLM: Large Language Models Are Effective Temporal Learners",
    booktitle="Computer Vision -- ECCV 2024",
    year="2025",
    publisher="Springer Nature Switzerland",
    address="Cham",
    pages="1--18",
    isbn="978-3-031-72998-0"
}

@misc{Munasinghe_2023_arXiv_PGVideoLLaVA,
      title={PG-Video-LLaVA: Pixel Grounding Large Video-Language Models}, 
      author={Shehan Munasinghe and Rusiru Thushara and Muhammad Maaz and Hanoona Abdul Rasheed and Salman Khan and Mubarak Shah and Fahad Khan},
      year={2023},
      eprint={2311.13435},
      archivePrefix={arXiv},
      primaryClass={cs.CV},
      url={https://arxiv.org/abs/2311.13435}, 
}

@misc{Maaz_2024_arXiv_VideoChatGPT_Temporal,
      title={Video-ChatGPT: Towards Detailed Video Understanding via Large Vision and Language Models}, 
      author={Muhammad Maaz and Hanoona Rasheed and Salman Khan and Fahad Shahbaz Khan},
      year={2024},
      eprint={2306.05424},
      archivePrefix={arXiv},
      primaryClass={cs.CV},
      url={https://arxiv.org/abs/2306.05424}, 
}

@InProceedings{Campello_2013_Springer_HDBSCAN,
    author="Campello, Ricardo J. G. B. and Moulavi, Davoud and Sander, Joerg",
    editor="Pei, Jian and Tseng, Vincent S. and Cao, Longbing and Motoda, Hiroshi and Xu, Guandong",
    title="Density-Based Clustering Based on Hierarchical Density Estimates",
    booktitle="Advances in Knowledge Discovery and Data Mining",
    year="2013",
    publisher="Springer Berlin Heidelberg",
    address="Berlin, Heidelberg",
    pages="160--172",
    isbn="978-3-642-37456-2"
}

@InProceedings{Li_2025_CVPR_LLaVAST,
    author    = {Li, Hongyu and Chen, Jinyu and Wei, Ziyu and Huang, Shaofei and Hui, Tianrui and Gao, Jialin and Wei, Xiaoming and Liu, Si},
    title     = {LLaVA-ST: A Multimodal Large Language Model for Fine-Grained Spatial-Temporal Understanding},
    booktitle = {Proceedings of the IEEE/CVF Conference on Computer Vision and Pattern Recognition (CVPR)},
    month     = {June},
    year      = {2025},
    pages     = {8592-8603}
}

@misc{Mao_2025_arXiv_SpatialLM,
      title={SpatialLM: Training Large Language Models for Structured Indoor Modeling}, 
      author={Yongsen Mao and Junhao Zhong and Chuan Fang and Jia Zheng and Rui Tang and Hao Zhu and Ping Tan and Zihan Zhou},
      year={2025},
      eprint={2506.07491},
      archivePrefix={arXiv},
      primaryClass={cs.CV},
      url={https://arxiv.org/abs/2506.07491}, 
}

@misc{He_2024_arXiv_UniMOV3D,
      title={UniM-OV3D: Uni-Modality Open-Vocabulary 3D Scene Understanding with Fine-Grained Feature Representation}, 
      author={Qingdong He and Jinlong Peng and Zhengkai Jiang and Kai Wu and Xiaozhong Ji and Jiangning Zhang and Yabiao Wang and Chengjie Wang and Mingang Chen and Yunsheng Wu},
      year={2024},
      eprint={2401.11395},
      archivePrefix={arXiv},
      primaryClass={cs.CV},
      url={https://arxiv.org/abs/2401.11395}, 
}

@misc{Hong_2023_arXiv_3dLLM,
      title={3D-LLM: Injecting the 3D World into Large Language Models}, 
      author={Yining Hong and Haoyu Zhen and Peihao Chen and Shuhong Zheng and Yilun Du and Zhenfang Chen and Chuang Gan},
      year={2023},
      eprint={2307.12981},
      archivePrefix={arXiv},
      primaryClass={cs.CV},
      url={https://arxiv.org/abs/2307.12981}, 
}

@InProceedings{Yang_2022_CVPR_TubeDETR,
    author    = {Yang, Antoine and Miech, Antoine and Sivic, Josef and Laptev, Ivan and Schmid, Cordelia},
    title     = {TubeDETR: Spatio-Temporal Video Grounding With Transformers},
    booktitle = {Proceedings of the IEEE/CVF Conference on Computer Vision and Pattern Recognition (CVPR)},
    month     = {June},
    year      = {2022},
    pages     = {16442-16453}
}

@misc{Wang_2025_arXiv_SpaceVLLM,
      title={SpaceVLLM: Endowing Multimodal Large Language Model with Spatio-Temporal Video Grounding Capability}, 
      author={Jiankang Wang and Zhihan Zhang and Zhihang Liu and Yang Li and Jiannan Ge and Hongtao Xie and Yongdong Zhang},
      year={2025},
      eprint={2503.13983},
      archivePrefix={arXiv},
      primaryClass={cs.CV},
      url={https://arxiv.org/abs/2503.13983}, 
}

@ARTICLE{Tang_2022_IEEE_HCSTVG,
  author={Tang, Zongheng and Liao, Yue and Liu, Si and Li, Guanbin and Jin, Xiaojie and Jiang, Hongxu and Yu, Qian and Xu, Dong},
  journal={IEEE Transactions on Circuits and Systems for Video Technology}, 
  title={Human-Centric Spatio-Temporal Video Grounding With Visual Transformers}, 
  year={2022},
  volume={32},
  number={12},
  pages={8238-8249},
  doi={10.1109/TCSVT.2021.3085907}
}

@InProceedings{Zhang_2020_CVPR_VidSTG,
author = {Zhang, Zhu and Zhao, Zhou and Zhao, Yang and Wang, Qi and Liu, Huasheng and Gao, Lianli},
title = {Where Does It Exist: Spatio-Temporal Video Grounding for Multi-Form Sentences},
booktitle = {Proceedings of the IEEE/CVF Conference on Computer Vision and Pattern Recognition (CVPR)},
month = {June},
year = {2020}
}

@InProceedings{Li_2024_CVPR_VideoChat2_Temporal,
    author    = {Li, Kunchang and Wang, Yali and He, Yinan and Li, Yizhuo and Wang, Yi and Liu, Yi and Wang, Zun and Xu, Jilan and Chen, Guo and Luo, Ping and Wang, Limin and Qiao, Yu},
    title     = {MVBench: A Comprehensive Multi-modal Video Understanding Benchmark},
    booktitle = {Proceedings of the IEEE/CVF Conference on Computer Vision and Pattern Recognition (CVPR)},
    month     = {06},
    year      = {2024},
    pages     = {22195-22206}
}

@InProceedings{He_2024_CVPR_MA-LLM_Temporal,
    author    = {He, Bo and Li, Hengduo and Jang, Young Kyun and Jia, Menglin and Cao, Xuefei and Shah, Ashish and Shrivastava, Abhinav and Lim, Ser-Nam},
    title     = {MA-LMM: Memory-Augmented Large Multimodal Model for Long-Term Video Understanding},
    booktitle = {Proceedings of the IEEE/CVF Conference on Computer Vision and Pattern Recognition (CVPR)},
    month     = {06},
    year      = {2024},
    pages     = {13504-13514}
}

@InProceedings{Li_2025_ECCV_LLaMA-VID_Temporal,
    author="Li, Yanwei and Wang, Chengyao and Jia, Jiaya",
    editor="Leonardis, Ale{\v{s}} and Ricci, Elisa and Roth, Stefan and Russakovsky, Olga and Sattler, Torsten and Varol, G{\"u}l",
    title="LLaMA-VID: An Image is Worth 2 Tokens in Large Language Models",
    booktitle="Computer Vision -- ECCV 2024",
    year="2025",
    publisher="Springer Nature Switzerland",
    address="Cham",
    pages="323--340",
    isbn="978-3-031-72952-2"
}

@misc{Ravi_2024_arXiv_SAM2,
      title={SAM 2: Segment Anything in Images and Videos}, 
      author={Nikhila Ravi and Valentin Gabeur and Yuan-Ting Hu and Ronghang Hu and Chaitanya Ryali and Tengyu Ma and Haitham Khedr and Roman Rädle and Chloe Rolland and Laura Gustafson and Eric Mintun and Junting Pan and Kalyan Vasudev Alwala and Nicolas Carion and Chao-Yuan Wu and Ross Girshick and Piotr Dollár and Christoph Feichtenhofer},
      year={2024},
      eprint={2408.00714},
      archivePrefix={arXiv},
      primaryClass={cs.CV},
      url={https://arxiv.org/abs/2408.00714}, 
}

@misc{Zhu_2025_arXiv_LLaVA-3D,
      title={LLaVA-3D: A Simple yet Effective Pathway to Empowering LMMs with 3D-awareness}, 
      author={Chenming Zhu and Tai Wang and Wenwei Zhang and Jiangmiao Pang and Xihui Liu},
      year={2025},
      eprint={2409.18125},
      archivePrefix={arXiv},
      primaryClass={cs.CV},
      url={https://arxiv.org/abs/2409.18125}, 
}

@InProceedings{Zhang_2022_CVPR_PointCLIP,
    author    = {Zhang, Renrui and Guo, Ziyu and Zhang, Wei and Li, Kunchang and Miao, Xupeng and Cui, Bin and Qiao, Yu and Gao, Peng and Li, Hongsheng},
    title     = {PointCLIP: Point Cloud Understanding by CLIP},
    booktitle = {Proceedings of the IEEE/CVF Conference on Computer Vision and Pattern Recognition (CVPR)},
    month     = {June},
    year      = {2022},
    pages     = {8552-8562}
}

@InProceedings{Zhu_2023_ICCVPointCLIPv2,
    author    = {Zhu, Xiangyang and Zhang, Renrui and He, Bowei and Guo, Ziyu and Zeng, Ziyao and Qin, Zipeng and Zhang, Shanghang and Gao, Peng},
    title     = {PointCLIP V2: Prompting CLIP and GPT for Powerful 3D Open-world Learning},
    booktitle = {Proceedings of the IEEE/CVF International Conference on Computer Vision (ICCV)},
    month     = {October},
    year      = {2023},
    pages     = {2639-2650}
}

@InProceedings{Qi_2024_CVPR_GPT4Point,
    author    = {Qi, Zhangyang and Fang, Ye and Sun, Zeyi and Wu, Xiaoyang and Wu, Tong and Wang, Jiaqi and Lin, Dahua and Zhao, Hengshuang},
    title     = {GPT4Point: A Unified Framework for Point-Language Understanding and Generation},
    booktitle = {Proceedings of the IEEE/CVF Conference on Computer Vision and Pattern Recognition (CVPR)},
    month     = {June},
    year      = {2024},
    pages     = {26417-26427}
}

@misc{Chen_2024_arXiv_Grounded3DLLM,
      title={Grounded 3D-LLM with Referent Tokens}, 
      author={Yilun Chen and Shuai Yang and Haifeng Huang and Tai Wang and Runsen Xu and Ruiyuan Lyu and Dahua Lin and Jiangmiao Pang},
      year={2024},
      eprint={2405.10370},
      archivePrefix={arXiv},
      primaryClass={cs.CV},
      url={https://arxiv.org/abs/2405.10370}, 
}

@misc{Deng_2025_arXiv_3D-LLaVA,
      title={3D-LLaVA: Towards Generalist 3D LMMs with Omni Superpoint Transformer}, 
      author={Jiajun Deng and Tianyu He and Li Jiang and Tianyu Wang and Feras Dayoub and Ian Reid},
      year={2025},
      eprint={2501.01163},
      archivePrefix={arXiv},
      primaryClass={cs.CV},
      url={https://arxiv.org/abs/2501.01163}, 
}

@InProceedings{Zhu_2025_ECCV_ScanReason,
    author="Zhu, Chenming and Wang, Tai and Zhang, Wenwei and Chen, Kai and Liu, Xihui",
    editor="Leonardis, Ale{\v{s}} and Ricci, Elisa and Roth, Stefan and Russakovsky, Olga and Sattler, Torsten and Varol, G{\"u}l",
    title="ScanReason: Empowering 3D Visual Grounding with Reasoning Capabilities",
    booktitle="Computer Vision -- ECCV 2024",
    year="2025",
    publisher="Springer Nature Switzerland",
    address="Cham",
    pages="151--168",
    isbn="978-3-031-73242-3"
}

@ARTICLE{Ji_2025_IEEE_JM3D,
  author={Ji, Jiayi and Wang, Haowei and Wu, Changli and Ma, Yiwei and Sun, Xiaoshuai and Ji, Rongrong},
  journal={IEEE Transactions on Pattern Analysis and Machine Intelligence}, 
  title={JM3D \& JM3D-LLM: Elevating 3D Representation With Joint Multi-Modal Cues}, 
  year={2025},
  volume={47},
  number={4},
  pages={2475-2492},
  doi={10.1109/TPAMI.2024.3523675}
}

@misc{Liu_2024_arXiv_Uni3D-LLM,
      title={Uni3D-LLM: Unifying Point Cloud Perception, Generation and Editing with Large Language Models}, 
      author={Dingning Liu and Xiaoshui Huang and Yuenan Hou and Zhihui Wang and Zhenfei Yin and Yongshun Gong and Peng Gao and Wanli Ouyang},
      year={2024},
      eprint={2402.03327},
      archivePrefix={arXiv},
      primaryClass={cs.CV},
      url={https://arxiv.org/abs/2402.03327}, 
}

@misc{Yang_2023_arXiv_LiDAR-LLM,
      title={LiDAR-LLM: Exploring the Potential of Large Language Models for 3D LiDAR Understanding}, 
      author={Senqiao Yang and Jiaming Liu and Ray Zhang and Mingjie Pan and Zoey Guo and Xiaoqi Li and Zehui Chen and Peng Gao and Yandong Guo and Shanghang Zhang},
      year={2023},
      eprint={2312.14074},
      archivePrefix={arXiv},
      primaryClass={cs.CV},
      url={https://arxiv.org/abs/2312.14074}, 
}

@misc{Guo_2023_arXiv_Point-LLM,
      title={Point-Bind \& Point-LLM: Aligning Point Cloud with Multi-modality for 3D Understanding, Generation, and Instruction Following}, 
      author={Ziyu Guo and Renrui Zhang and Xiangyang Zhu and Yiwen Tang and Xianzheng Ma and Jiaming Han and Kexin Chen and Peng Gao and Xianzhi Li and Hongsheng Li and Pheng-Ann Heng},
      year={2023},
      eprint={2309.00615},
      archivePrefix={arXiv},
      primaryClass={cs.CV},
      url={https://arxiv.org/abs/2309.00615}, 
}

@article{Qian_2024_AAAI_NuScenes-QA, 
    title={NuScenes-QA: A Multi-Modal Visual Question Answering Benchmark for Autonomous Driving Scenario}, 
    volume={38}, 
    url={https://ojs.aaai.org/index.php/AAAI/article/view/28253}, 
    DOI={10.1609/aaai.v38i5.28253}, 
    number={5}, 
    journal={Proceedings of the AAAI Conference on Artificial Intelligence}, 
    author={Qian, Tianwen and Chen, Jingjing and Zhuo, Linhai and Jiao, Yang and Jiang, Yu-Gang}, 
    year={2024}, 
    month={Mar.}, 
    pages={4542-4550} 
}

@misc{Hwang_2024_arXiv_EMMA,
      title={EMMA: End-to-End Multimodal Model for Autonomous Driving}, 
      author={Jyh-Jing Hwang and Runsheng Xu and Hubert Lin and Wei-Chih Hung and Jingwei Ji and Kristy Choi and Di Huang and Tong He and Paul Covington and Benjamin Sapp and Yin Zhou and James Guo and Dragomir Anguelov and Mingxing Tan},
      year={2024},
      eprint={2410.23262},
      archivePrefix={arXiv},
      primaryClass={cs.CV},
      url={https://arxiv.org/abs/2410.23262}, 
}

@InProceedings{Zhou_2025_ECCV_ELM,
    author="Zhou, Yunsong and Huang, Linyan and Bu, Qingwen and Zeng, Jia and Li, Tianyu and Qiu, Hang and Zhu, Hongzi and Guo, Minyi and Qiao, Yu and Li, Hongyang",
    editor="Leonardis, Ale{\v{s}} and Ricci, Elisa and Roth, Stefan and Russakovsky, Olga and Sattler, Torsten and Varol, G{\"u}l",
    title="Embodied Understanding of Driving Scenarios",
    booktitle="Computer Vision -- ECCV 2024",
    year="2025",
    publisher="Springer Nature Switzerland",
    address="Cham",
    pages="129--148",
    isbn="978-3-031-73033-7"
}

@ARTICLE{Cui_IEEE_2024_DriveAsYouSay_LLM,
  author={Cui, Can and Ma, Yunsheng and Cao, Xu and Ye, Wenqian and Wang, Ziran},
  journal={IEEE Intelligent Transportation Systems Magazine}, 
  title={Receive, Reason, and React: Drive as You Say, With Large Language Models in Autonomous Vehicles}, 
  year={2024},
  volume={16},
  number={4},
  pages={81-94},
  doi={10.1109/MITS.2024.3381793}
}

@InProceedings{Cui_2024_WACV_DriveAsYouSpeak_LLM,
    author    = {Cui, Can and Ma, Yunsheng and Cao, Xu and Ye, Wenqian and Wang, Ziran},
    title     = {Drive As You Speak: Enabling Human-Like Interaction With Large Language Models in Autonomous Vehicles},
    booktitle = {Proceedings of the IEEE/CVF Winter Conference on Applications of Computer Vision (WACV) Workshops},
    month     = {January},
    year      = {2024},
    pages     = {902-909}
}

@misc{Sohn_2025_arXiv_Framework4C,
      title={A Framework for a Capability-driven Evaluation of Scenario Understanding for Multimodal Large Language Models in Autonomous Driving}, 
      author={Tin Stribor Sohn and Philipp Reis and Maximilian Dillitzer and Johannes Bach and Jason J. Corso and Eric Sax},
      year={2025},
      eprint={2503.11400},
      archivePrefix={arXiv},
      primaryClass={cs.CV},
      url={https://arxiv.org/abs/2503.11400}, 
}

@misc{Google_2025_Gemma3,
      title={Gemma 3 Technical Report}, 
      author={{Gemma Team}},
      year={2025},
      eprint={2503.19786},
      archivePrefix={arXiv},
      primaryClass={cs.CL},
      url={https://arxiv.org/abs/2503.19786}, 
}

@article{Huhn_1955_HungarianMatching,
  title={The Hungarian method for the assignment problem},
  author={Kuhn, Harold W},
  journal={Naval research logistics quarterly},
  volume={2},
  number={1-2},
  pages={83--97},
  year={1955},
  publisher={Wiley Online Library}
}

@misc{Guadagnino_2025_arXiv_kissSLAM,
      title={KISS-SLAM: A Simple, Robust, and Accurate 3D LiDAR SLAM System With Enhanced Generalization Capabilities}, 
      author={Tiziano Guadagnino and Benedikt Mersch and Saurabh Gupta and Ignacio Vizzo and Giorgio Grisetti and Cyrill Stachniss},
      year={2025},
      eprint={2503.12660},
      archivePrefix={arXiv},
      primaryClass={cs.RO},
      url={https://arxiv.org/abs/2503.12660}, 
}

@misc{Mao_2023_arXiv_GPT-Driver,
      title={GPT-Driver: Learning to Drive with GPT}, 
      author={Jiageng Mao and Yuxi Qian and Junjie Ye and Hang Zhao and Yue Wang},
      year={2023},
      eprint={2310.01415},
      archivePrefix={arXiv},
      primaryClass={cs.CV},
      url={https://arxiv.org/abs/2310.01415}, 
}

@ARTICLE{Xu_2024_IEEE_DriveGPT4,
  author={Xu, Zhenhua and Zhang, Yujia and Xie, Enze and Zhao, Zhen and Guo, Yong and Wong, Kwan-Yee K. and Li, Zhenguo and Zhao, Hengshuang},
  journal={IEEE Robotics and Automation Letters}, 
  title={DriveGPT4: Interpretable End-to-End Autonomous Driving Via Large Language Model}, 
  year={2024},
  volume={9},
  number={10},
  pages={8186-8193},
  doi={10.1109/LRA.2024.3440097}
}

@misc{Wang_2024_arXiv_TemporalReasoning,
      title={TRAM: Benchmarking Temporal Reasoning for Large Language Models}, 
      author={Yuqing Wang and Yun Zhao},
      year={2024},
      eprint={2310.00835},
      archivePrefix={arXiv},
      primaryClass={cs.CL},
      url={https://arxiv.org/abs/2310.00835}, 
}

@misc{Chu_2024_arXiv_TemporalReasoning,
      title={TimeBench: A Comprehensive Evaluation of Temporal Reasoning Abilities in Large Language Models}, 
      author={Zheng Chu and Jingchang Chen and Qianglong Chen and Weijiang Yu and Haotian Wang and Ming Liu and Bing Qin},
      year={2024},
      eprint={2311.17667},
      archivePrefix={arXiv},
      primaryClass={cs.CL},
      url={https://arxiv.org/abs/2311.17667}, 
}

@InProceedings{Ranasinghe_2024_CVPR_LLMSpatial,
    author    = {Ranasinghe, Kanchana and Shukla, Satya Narayan and Poursaeed, Omid and Ryoo, Michael S. and Lin, Tsung-Yu},
    title     = {Learning to Localize Objects Improves Spatial Reasoning in Visual-LLMs},
    booktitle = {Proceedings of the IEEE/CVF Conference on Computer Vision and Pattern Recognition (CVPR)},
    month     = {June},
    year      = {2024},
    pages     = {12977-12987}
}

@misc{Wang_2024_arXiv_LLMSpatial,
      title={Is A Picture Worth A Thousand Words? Delving Into Spatial Reasoning for Vision Language Models}, 
      author={Jiayu Wang and Yifei Ming and Zhenmei Shi and Vibhav Vineet and Xin Wang and Yixuan Li and Neel Joshi},
      year={2024},
      eprint={2406.14852},
      archivePrefix={arXiv},
      primaryClass={cs.CV},
      url={https://arxiv.org/abs/2406.14852}, 
}

@misc{Zhang_2024_arXiv_LLMSpatial,
      title={CounterCurate: Enhancing Physical and Semantic Visio-Linguistic Compositional Reasoning via Counterfactual Examples}, 
      author={Jianrui Zhang and Mu Cai and Tengyang Xie and Yong Jae Lee},
      year={2024},
      eprint={2402.13254},
      archivePrefix={arXiv},
      primaryClass={cs.CV},
      url={https://arxiv.org/abs/2402.13254}, 
}

@misc{Jiao_2025_arXiv_LaVidaDrive,
      title={LaVida Drive: Vision-Text Interaction VLM for Autonomous Driving with Token Selection, Recovery and Enhancement}, 
      author={Siwen Jiao and Yangyi Fang and Baoyun Peng and Wangqun Chen and Bharadwaj Veeravalli},
      year={2025},
      eprint={2411.12980},
      archivePrefix={arXiv},
      primaryClass={cs.CV},
      url={https://arxiv.org/abs/2411.12980}, 
}

@InProceedings{Ding_2024_CVPR_BEV-InMLLM,
    author    = {Ding, Xinpeng and Han, Jianhua and Xu, Hang and Liang, Xiaodan and Zhang, Wei and Li, Xiaomeng},
    title     = {Holistic Autonomous Driving Understanding by Bird's-Eye-View Injected Multi-Modal Large Models},
    booktitle = {Proceedings of the IEEE/CVF Conference on Computer Vision and Pattern Recognition (CVPR)},
    month     = {06},
    year      = {2024},
    pages     = {13668-13677}
}

@misc{Zha_2025_arXiv_SurveySpatialLLMs,
      title={How to Enable LLM with 3D Capacity? A Survey of Spatial Reasoning in LLM}, 
      author={Jirong Zha and Yuxuan Fan and Xiao Yang and Chen Gao and Xinlei Chen},
      year={2025},
      eprint={2504.05786},
      archivePrefix={arXiv},
      primaryClass={cs.CV},
      url={https://arxiv.org/abs/2504.05786}, 
}
